\documentclass[letterpaper,journal]{IEEEtran}
\usepackage{amsmath,amsfonts}
\usepackage{algorithmic}
\usepackage{algorithm}
\usepackage{array}
\usepackage[caption=false,font=normalsize,labelfont=sf,textfont=sf]{subfig}
\usepackage{textcomp}
\usepackage{stfloats}
\usepackage{url}
\usepackage{hyperref}
\usepackage{verbatim}
\usepackage{graphicx}
\usepackage{cite}
\usepackage{xcolor}
\usepackage{subfig}
\usepackage[utf8]{inputenc}
\usepackage[T1]{fontenc}
\usepackage{booktabs}
\definecolor{first}{HTML}{DC2626}
\definecolor{second}{HTML}{4A90E2}
\definecolor{third}{HTML}{35B779}
\definecolor{resblue}{HTML}{2E86AB}
\definecolor{resred}{HTML}{C73E1D}
\definecolor{resgreen}{HTML}{0F7B0F}

\definecolor{darkgreen}{rgb}{0.0, 0.6, 0.0}
\definecolor{coolblue}{rgb}{0.0, 0.5, 0.75}

\hypersetup{
    colorlinks=true,
    linkcolor=red,
    urlcolor=coolblue,    
    citecolor=darkgreen,
}

\newcommand{\etal}{~{\it et al.} }

\begin{document}

\title{Local Virtual Nodes for Alleviating \\Over-Squashing in Graph Neural Networks}

\author{Tuğrul Hasan Karabulut and İnci M. Baytaş,~\IEEEmembership{~IEEE,}
	\thanks{Tuğrul Hasan Karabulut and İnci M. Baytaş are with 
the Computer Engineering Department, Bo\u{g}azi\c{c}i University, Istanbul, Turkey (e-mail: tugrul.karabulut@std.bogazici.edu.tr, inci.baytas@bogazici.edu.tr).}
	\thanks{This paper was produced by the IEEE Publication Technology Group. They are in Piscataway, NJ.}
	\thanks{Manuscript received April 19, 2021; revised August 16, 2021.}}

\markboth{Journal of \LaTeX\ Class Files,~Vol.~14, No.~8, August~2021}%
{Shell \MakeLowercase{\textit{et al.}}: A Sample Article Using IEEEtran.cls for IEEE Journals}


\maketitle

\begin{abstract}
	Over-squashing is a challenge in training graph neural networks for tasks
	involving long-range dependencies. In such tasks, a GNN's receptive field
	should be large enough to enable communication between distant nodes.
	However, gathering information from a wide range of neighborhoods and
	squashing its content into fixed-size node representations makes
	message-passing vulnerable to bottlenecks. Graph rewiring and adding virtual
	nodes are commonly studied remedies that create additional pathways around
	bottlenecks to mitigate over-squashing. However, these techniques alter the
	input graph's global topology and disrupt the domain knowledge encoded in
	the original graph structure, both of which could be essential to specific
	tasks and domains. This study presents Local Virtual Nodes (LVN) with
	trainable embeddings to alleviate the effects of over-squashing without
	significantly corrupting the global structure of the input graph. The
	position of the LVNs is determined by the node centrality, which indicates
	the existence of potential bottlenecks. Thus, the proposed approach aims to
	improve the connectivity in the regions with likely bottlenecks.
	Furthermore, trainable LVN embeddings shared across selected central regions
	facilitate communication between distant nodes without adding more layers.
	Extensive experiments on benchmark datasets demonstrate that LVNs can enhance structural connectivity and
	significantly improve performance on graph and node classification tasks.
	The code can be found
	at~\href{https://github.com/ALLab-Boun/LVN/}{https://github.com/ALLab-Boun/LVN/}.
\end{abstract}



\section{Introduction}

\IEEEPARstart{G}{raph} Neural Networks (GNNs) have become a standard for representation learning
in tasks involving non-Euclidean data. They can handle arbitrary graph
topologies without any prior assumptions about their structure. Therefore, the
GNNs are integral components in applications of social
networks~\cite{borisyuk2024lignn}, traffic
networks~\cite{qi2023graph}, and molecular
graphs~\cite{zhang2024gnngo3d}. The flexibility of GNNs and their
applicability to various domains mainly stem from the message-passing paradigm,
an efficient operation that can handle diverse graph structures at a massive
scale~\cite{gilmer,hamilton2017inductive}. In message-passing, node
representations are exchanged as messages among the adjacent nodes. The messages
from neighboring nodes are later aggregated with a permutation-invariant
operation to update the node representations~\cite{gilmer}.

Multiple message-passing layers can be stacked to
improve the expressivity of GNNs. 
Although there is no theoretical upper limit to the
number of layers required by the downstream tasks, GNNs suffer from two of the
most commonly studied issues with increasing number of message-passing layers in
practice, over-smoothing~\cite{li2018deeper}, and
over-squashing~\cite{alon2021on}. Adding message-passing layers enlarges the
receptive field of the GNN. Incorporating the information from distant nodes to
update the node representations might be helpful for long-range applications.
However, aggregating information from a wide range of neighborhoods can also
result in almost indistinguishable node representations, known as
over-smoothing~\cite{li2018deeper}. On the other hand, graphs with severe
bottlenecks might prevent the information flow from distant nodes, known as
over-squashing.

Most of the traditional GNN architectures, such as Graph Convolutional Network
(GCN) and GraphSAGE~\cite{hamilton2017inductive}, suffer from over-smoothing,
which hurts the downstream task performance~\cite{li2018deeper}. Many studies have
been proposed to combat over-smoothing, ranging from sophisticated
methods~\cite{rusch2022graph,lee2023towards,karabulut2024channel} to minor
architectural modifications~\cite{xu2018jumping,chen2020simple}. On the other
hand, over-squashing remains a more challenging issue in GNN literature as it
prevents practitioners from applying GNNs to problems with a large radius. This
limitation forces studies to resort to computationally complex methods such as
graph transformers~\cite{ying2021do,ma2023graph,muller2024attending}. Therefore,
developing effective methods that enable long-range learning in GNNs represents
a crucial need for more computationally efficient alternatives.

Each successive GNN layer causes an exponential increase of information encoded
in the node representations~\cite{alon2021on}. Squashing excessive
information into fixed-size representations poses a challenge when the graph has
bottleneck regions blocking the information exchange between distant nodes.
Therefore, GNN architectures with multiple message-passing layers and
fixed-length node representations are prone to over-squashing, which prevents
effective learning on graphs. Recent studies tackle over-squashing
from a topology or an architecture perspective. The methods proposing
topological solutions, namely \textit{graph rewiring}, alter the graph topology
by adding new edges that create new pathways for information propagation, and optionally remove redundant edges to reduce the
computational
overhead~\cite{topping2022understanding,karhadkar2023fosr,nguyen2023revisiting,black2023understanding}.

Topological approaches can be further classified into spectral and spatial
methods. Spectral graph rewiring aims to optimize a metric
based on the graph Laplacian matrix~\cite{karhadkar2023fosr,arnaiz2022diffwire}.
Spatial rewiring, on the other hand, is used to improve the local connectivity
of the graph~\cite{gabrielsson2023rewiring,barbero2024localityaware}. Although
graph rewiring helps eliminate the structural reasons for bottlenecks in the
graphs, it disrupts the valuable domain knowledge represented in the graph
structure. Furthermore, finding the bottlenecks in graphs is computationally
infeasible in large graphs, as it requires eigendecomposition of the graph
Laplacian or graph curvature measures. In addition, graph rewiring might create
ambiguity for edge-level tasks since new edges are introduced while some
existing ones are removed.

Besides modifying the graph structure with rewiring, studies show that
over-squashing can also be mitigated by increasing the representation capacity
of the GNNs. Di Giovanni\etal\cite{di2023over} theoretically validated that
increasing the width of GNNs by defining an influence measure based on the norm
of the Jacobian of node representations from the input and the output layers. The
authors proved that the influence measure's upper bound depends on the width of
the GNN. However, expanding the network width increases the computational
complexity and risk of overfitting~\cite{di2023over}. Therefore, it is essential
to develop ways to increase the capacity of the graph representation
learning to combat over-squashing without tampering with the domain and the task. Choi\etal proposed Expanded Width-Aware
Message-Passing (PANDA), a recent width-expansion method that increases the
width of central node embeddings~\cite{choi2024panda}. However, PANDA incorporates
additional layers and modifies the GNN architecture to handle message-passing
between nodes of different dimensions, which introduces additional
parameters~\cite{choi2024panda}.

To mitigate over-squashing, this paper proposes to modify the node connectivity locally without distorting the graph's global topology. We introduce \textit{Local Virtual Nodes} (LVN), which support high-centrality nodes in the graph. By adding LVNs to central nodes, we aim to increase the information capacity and facilitate message-passing in regions likely to have bottlenecks. Furthermore, we seek to achieve communication among distant nodes by assigning trainable embeddings to LVNs shared across selected central regions in the graph. Thus, the LVNs offer two key benefits: creating additional pathways for information flow from dense regions to sparse regions, and enhancing feature
representations that operate independently of the GNN's standard receptive field through the shared trainable embeddings. Training GNNs with LVNs differs from graph
rewiring and global virtual node techniques in preserving the graph's global structure and the domain knowledge represented by the original connections. Therefore, we do not connect distant nodes to obtain long-range
connections like spectral rewiring methods. Unlike spatial rewiring, we avoid connecting unrelated nodes to improve connectivity. In addition, the LVNs do not cause a substantial global topological change, unlike adding global virtual nodes that immensely increase the number of edges in large-scale settings. The contributions of the study are highlighted as:
\begin{itemize}
	\item We propose Local Virtual Nodes (LVNs) for expanding the representation capacity of a graph's central regions. Each central node is augmented with a group of LVNs.
	\item LVNs increase the number of paths through central regions, enhancing connectivity and message-passing.
	\item An LVN is equipped with a trainable embedding. LVNs in the same group do not share the trainable embeddings. However, embeddings are shared across LVN groups associated with different central regions in the graph.

	\item Sharing LVN embeddings across groups allows different central regions to learn complementary representations within a unified feature space, enabling long-range communication during message-passing.

	\item The proposed approach is compatible with any existing GNN architecture in the literature.
\end{itemize}
Extensive experiments are conducted with various benchmark datasets for graph and node classification tasks. We analyze the effectiveness of LVNs in improving the connectivity of the graph based on various structural metrics. The experimental results show that adding LVNs with trainable embeddings can mitigate the effects of over-squashing on connectivity and the performance of downstream tasks. The proposed method outperforms well-known graph rewiring methods in the literature by a large margin.

\section{Related Work}

The primary cause of over-squashing is attributed to graph
topology~\cite{di2023over, giovanni2024how}. Therefore, the field has been more
focused on separating the input graph and the graph fed into the GNN,
i.e., the computational graph. Alon and Yahav~\cite{alon2021on} were among the
first to propose such a solution that uses a complete graph in the GNN's last layer. Graph transformers similarly decouple the computational and the input
graph by treating the input as a complete graph while learning the structure
with attention and the positional
encodings~\cite{ying2021do,ma2023graph,muller2024attending}. However, the
methods leveraging complete graphs have quadratic complexity regarding the
number of nodes, as opposed to linear complexity of message-passing with the raw
graph, and are infeasible to use in scenarios with scarce resources and large
graphs. Consequently, modifying the graph structure before training GNNs has become a favorable approach to mitigate over-squashing and improve overall graph connectivity. The term \textit{graph rewiring} refers to
such structural modification techniques in the literature.

Most graph rewiring approaches propose preprocessing steps before training.
Structural properties related to graph connectivity, such as curvature, effective resistance, and spectral gap, serve as guides to identify key parts
of the graphs that need
improvement~\cite{topping2022understanding,banerjee2022oversquashing,black2023understanding,karhadkar2023fosr,nguyen2023revisiting,barbero2024localityaware}.
Topping\etal introduced Stochastic Discrete Ricci Flow (SDRF), which depends on a
metric called Ricci curvature that quantifies to what extent an edge acts as a
bottleneck, and then adds edges around bottleneck edges while removing redundant
edges~\cite{topping2022understanding}. Another curvature-based rewiring
technique is Batch Ollivier-Ricci Flow (BORF), proposed by
Nguyen\etal~\cite{nguyen2023revisiting}. Unlike SDRF, BORF tackles both
over-smoothing and over-squashing using Ollivier-Ricci
curvature~\cite{nguyen2023revisiting}. In addition to curvature, spectral
metrics such as spectral gap and effective resistance have also been used to
identify regions to rewire in the
graph~\cite{banerjee2022oversquashing,black2023understanding,shen2024graph,karhadkar2023fosr}.
Black\etal introduced the Greedy Total Resistance (GTR) technique that rewires
the graph to optimize total effective resistance, a concept adapted from
electrical circuit theory to graph structures~\cite{black2023understanding}.
Greedy Local Edge Flip (G-RLEF) also rewires the graph by sampling node pairs
based on effective resistance and adding edges around the sampled node
pairs~\cite{banerjee2022oversquashing}. Graph Preprocessing with Effective
Resistance (GPER) combats over-squashing and over-smoothing by proposing
metrics to pinpoint redundant edges to remove and bottleneck regions to support
with additional edges~\cite{shen2024graph}. Spectral gap has also been utilized
in graph rewiring literature~\cite{karhadkar2023fosr}. Karhadkar\etal proposed
First-order Spectral Rewiring (FoSR) that adds edges to optimize the spectral
gap, a global metric that indicates the graph's
connectivity~\cite{karhadkar2023fosr}. Further, the authors proposed using
relational GNNs that handle the original and new edges
differently~\cite{karhadkar2023fosr}.

Some studies propose more conservative rewiring methods in terms of locality to avoid disrupting the inductive bias of the input graph. Diffusion Improves Graph
Learning (DIGL) is one of the seminal local rewiring techniques that learns a
graph diffusion matrix by combining powers of the adjacency matrix and feeds the sparsified diffusion matrix as the input graph to the
GNN~\cite{gasteiger2019diffusion}. Gabrielsson\etal suggest that applying local
rewiring to connect nodes within a certain distance, combined with graph
positional encoding, achieves significant performance improvement over the baseline
GNNs and graph transformers~\cite{gabrielsson2023rewiring}. Locality-Aware
Sequential Rewiring (LASER) is a more systematic local rewiring method that
preserves the graph's locality and sparsity~\cite{barbero2024localityaware}. On the other hand, some studies propose dynamically learning how to rewire the graph rather than
applying a step-by-step preprocessing
algorithm~\cite{arnaiz2022diffwire,qian2024probabilistically}. DiffWire is a dynamic and spectral rewiring approach that estimates a graph's commute time and
spectral gap with a neural network, and then rewires the graph based
on the estimated values~\cite{arnaiz2022diffwire}. Qian\etal proposed a
probabilistic rewiring framework that first generates scores for each edge, then
applies a differentiable sampling that selects edges to construct the rewired
graph~\cite{qian2024probabilistically}.

Lastly, another subfield in graph rewiring focuses on utilizing graph structures
known for their strong connectivity properties and the absence of bottlenecks.
These methods propose converting the given graph to a specific type of graph with which message-passing GNNs may work well. Expander, Cayley, and Delaunay graphs are known to have strong connectivity properties and are
used to circumvent structural limitations in the original
graph~\cite{deac2022expander, wilson2024cayley, attali2024delaunay}. The proposed local virtual nodes with trainable embeddings approach addresses some of the shortcomings of graph rewiring techniques.
Unlike graph rewiring, we do not significantly disrupt the graph's local and global topology. In addition, we do not break the domain knowledge represented by the
original edge set by adding and removing edges that might hurt downstream task performance.

\begin{figure}[t!]
	\centering
	\includegraphics[width=\columnwidth]{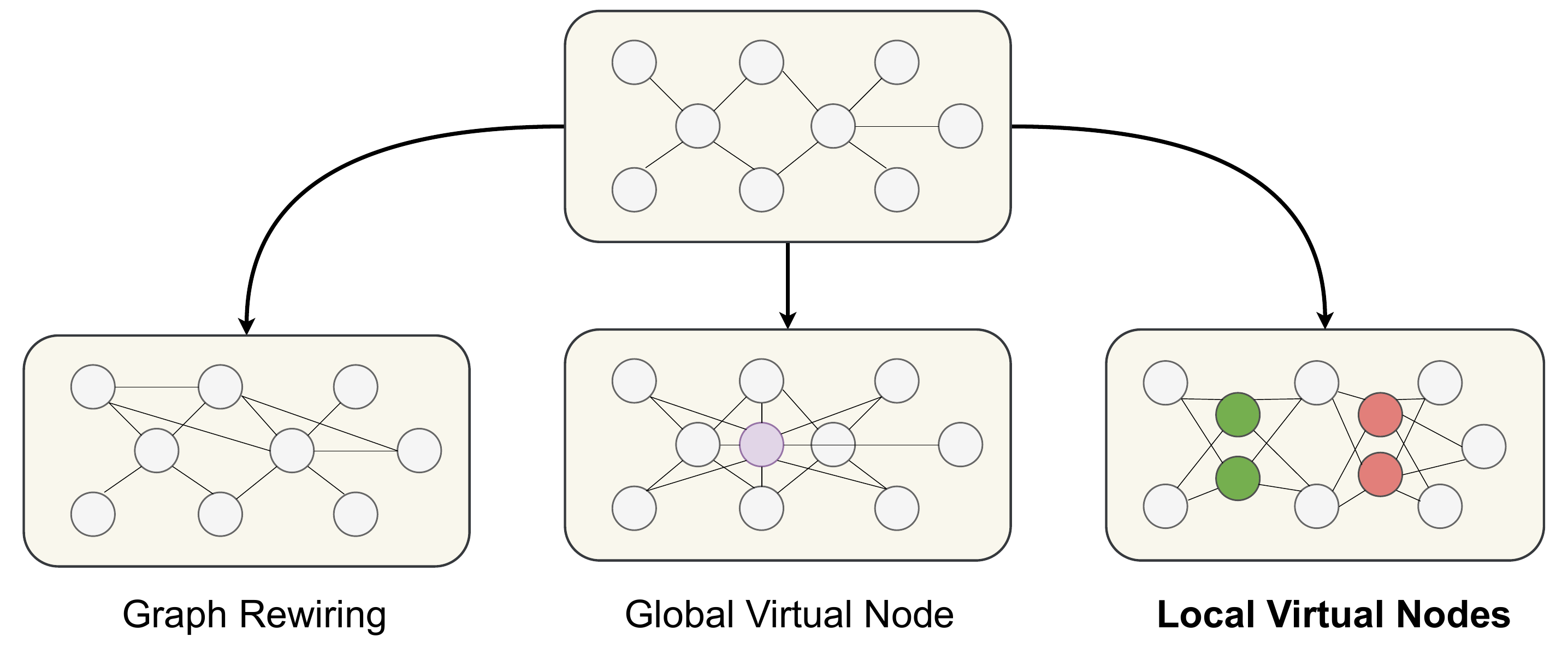}
	\caption{Comparison between graph rewiring, global virtual node, and local virtual nodes techniques.}
	\label{fig:methods_comparison}
\end{figure}

Virtual nodes are studied as an alternative that enables long-range
communication while avoiding the quadratic computational complexity of complete
graph methods. A virtual node is originally defined as a \textit{supernode} that
is connected to all other nodes~\cite{gilmer}. Cai\etal\cite{cai2023connection}
proved that a message-passing GNN with a virtual node and a large enough width
can simulate transformers. Further,
Southern\etal\cite{southern2025understanding} showed that a global virtual node
can help reduce the overall commute time, a metric computed using the graph's
spectrum. The virtual node approach is applied to various
tasks~\cite{hwang2022an,sestak2024vnegnn,zhang2024improving}. VN-EGNN extends
EGNN~\cite{satorras2021equivariant} with a set of global virtual nodes per graph
to solve the protein binding site identification task~\cite{sestak2024vnegnn}.
Concurrently, Zhang\etal employed global virtual nodes to improve learning on
large geometric graphs~\cite{zhang2024improving}. Hwang\etal proposed
incorporating virtual nodes for the link prediction task~\cite{hwang2022an}.
They include virtual nodes in the graph by clustering the graph and assigning
each cluster a virtual node. However, they did not focus on mitigating
over-squashing, but on analyzing the expressiveness under the Weisfeiler-Leman
test. Lastly, Qian\etal proposed incorporating a small number of virtual nodes
into the graph and probabilistically rewiring the graph by assigning exactly one
virtual node to each original node~\cite{qian2024probabilistic}. This approach
is similar to Hwang~{\it et al.}'s approach~\cite{hwang2022an}, but the
assignments are determined using a neural network rather than a
community-detection algorithm~\cite{qian2024probabilistic}. In this study, we
introduce strictly local virtual nodes (LVNs) instead of global virtual ones.
Moreover, the LVNs are considered a means to increase the representation
capacity of a central node rather than merely as additional nodes added to the
graph. Finally, the proposed globally shared trainable virtual node embeddings
further serve as a tool to extend the receptive field of GNNs without adding
extra message-passing layers. Figure~\ref{fig:methods_comparison} illustrates
the differences between the current structural approaches and our method. We can
observe that our approach improves connectivity without disrupting the graph's
locality.

Besides virtual nodes, expanding the width of some selected nodes to mitigate
over-squashing has also been explored in a framework named PANDA~\cite{choi2024panda}. In PANDA, the
feature dimensionality of a subset of nodes is increased without changing the
graph topology. More precisely, a set of central nodes is determined based on a
centrality measure. The selected nodes are assigned higher-dimensional
representations than the standard width considered in the architecture. Moreover, different message-passing
functions are applied for each pair of node types (standard-width and high-width
nodes)~\cite{choi2024panda}. Our framework offers several advantages over PANDA.
First, PANDA employs computationally intensive centrality measures (Betweenness,
Closeness, and Load~\cite{choi2024panda}) that require shortest path
calculations and run in quadratic
time~\cite{brandes2001faster,newman2001scientific}. In contrast, our approach
substantially improves the performance while using only computationally efficient
centrality measures suitable for large graphs. Further, PANDA requires
different message-passing functions between nodes of different
widths~\cite{choi2024panda}. Messages from low-width to high-width nodes use
weight matrices that increase feature dimensionality. In contrast, messages from
high-width to low-width nodes employ feature selection rather than dimensionality reduction~\cite{choi2024panda}. Conversely,
the proposed LVN framework integrates with any existing GNN without modifying the message-passing operations.


\section{Preliminaries}

This section presents the background and notation used in the paper.
We consider undirected graphs denoted by $G = (\mathcal{V}, \mathcal{E})$ without self-loops and edge weights, where $\mathcal{V}$ is the set of nodes and $\mathcal{E}$ is the set of edges.
The set of node indices $\{0,1,2,\ldots,N\}$ is denoted by $[N]$, where $N$ is the number of nodes in the graph. $G[S]$ represents the induced subgraph of $G$ for
the set $S \subseteq \mathcal{V}$, which includes all the edges betweeen nodes
in $S$. Alternatively, a graph is represented by its adjacency matrix $\mathbf{A} \in \mathbb{R}^{N \times
		N}$, and a diagonal degree matrix, $\mathbf{D}$, in which each entry is the node's degree. Neighborhood of a node $v \in
	\mathcal{V}$ is denoted by the set $\mathcal{N}(v)$. Each node has an input feature vector
$\mathbf{x} \in \mathbb{R}^F$, and all the node features of the graph form an input feature
matrix $\mathbf{X} \in \mathbb{R}^{N \times F}$, where $F$ is the input dimensionality. Meanwhile, the hidden dimensionality of a GNN's $l$-th hidden layer is denoted by $D$. The output of a hidden layer for node $v$ is referred to as $\mathbf{x}_v^{(l)} \in \mathbb{R}^D$. In node classification tasks, each node must have a label
$\mathbf{y} \in \{ 0, 1 \}^K$, where $K$ is the number of available classes in
the dataset. For graph classification tasks, we have a dataset defined as
$\mathcal{D} = \{ (G_1, y_1), (G_2, y_2), \dots, (G_n, y_n) \}$ with $n$ graphs
along with their corresponding labels.

\subsection{Message-Passing GNNs}

Message-passing constitutes the core operation in GNNs, which enables
structure-aware representation learning. The $l$-th layer of a GNN can be
denoted as
\begin{equation}
	\mathbf{m}_v^{(l)} = \text{agg}^{(l)} (\{  \text{msg}^{(l)} ( \mathbf{x}_v^{(l-1)}, \mathbf{x}_u^{(l-1)} ) \mid u \in \mathcal{N}(v)  \})
	\label{eq:mp1}
\end{equation}
\begin{equation}
	\mathbf{x}_v^{(l)} = \text{upd}^{(l)} ( \mathbf{x}_v^{(l-1)},  \mathbf{m}_v^{(l)}  )
	\label{eq:mp2}
\end{equation}
where $\mathbf{m}_v^{(l)}$ denotes the aggregated message vector that node $v$
received from its neighborhood with the help of $\text{msg}^{(l)}$ and
$\text{agg}^{(l)}$, the message function that outputs the individual message for
that neighbor and the aggregation function that forms an aggregated
permutation-invariant vector, respectively. Finally, the update function
$\text{upd}^{(l)}$ combines the existing node features and the aggregated
message to generate the new features for node $v$. Most prominent GNNs, such as Graph Convolutional Network
(GCN)~\cite{kipf2017semisupervised}, Graph Isomorphism Network
(GIN)~\cite{xu2018how}, and Graph Attention Network
(GAT)~\cite{velickovic2018graph}, can be characterized using the framework in
Equations~\ref{eq:mp1} and~\ref{eq:mp2}.

\subsection{Global Virtual Nodes}

Over-squashing occurs when the model cannot store abundant information
in a limited feature space. Over-squashing can be attributed either to the
graph's topology or the GNN's width~\cite{di2023over}. Increasing the width may help
for graphs where the structural issues are not severely pathological and the
problem radius is low. However, an excessive increase in width may bring about
challenges in generalization~\cite{di2023over}. Consequently, graph rewiring has
received more attention in the literature as a convenient preprocessing step,
allowing standard GNNs to be implemented without modification. Graph rewiring
techniques help mitigate over-squashing by connecting distant nodes with direct
edges, which would otherwise be required to communicate through the bottleneck
pathways. Di Giovanni\etal provide a theoretical framework proving that both
approaches could help combat over-squashing~\cite{di2023over}. The authors
utilize the Jacobian of a GNN's output layer with respect to the first
layer's input:
\begin{equation}
	\left\|\frac{\partial \mathbf{h}_{v}^{(l)}}{\partial
		\mathbf{h}_{u}^{(0)}}\right\|_{1} \leq \underbrace{(c_\sigma
		wp)^l}_{\text{model}} \ \underbrace{(\mathbf{S}^{l})_{vu}}_{\text{topology}}
	\label{eq:sensitivity}
\end{equation}
where $c_\sigma$ is the Lipschitz constant of the activation function $\sigma$,
$w$ is the maximum value of all the weight matrices, $p$ is the model's width,
and $\mathbf{S}$ is the graph shift operator defined based on the adjacency
matrix, $\mathbf{A}$, and used for message-passing. For example, GCN's graph
shift operator is $\mathbf{D}^{-1/2} \mathbf{A} \mathbf{D}^{-1/2}$.

The Equation~(\ref{eq:sensitivity}) (Theorem 3.2 in~\cite{di2023over})
demonstrates that the sensitivity of a GNN's output with respect to its input is
dependent on both the model parameters, such as the activation function, and
network width, as well as the underlying graph topology. In the Equation~(\ref{eq:sensitivity}), $c_\sigma$ is derived from the
common activation function choices (ReLU, sigmoid, etc.), and the weights are
determined during training and are usually kept from exploding with
regularization. Thus, the network width $p$ and the graph shift operator
$\mathbf{S}$, which depends on the topology, remains as the configurable
parameters to combat over-squashing.

Some studies suggest global virtual nodes (GVN) to combat over-squashing. Adding GVN bridges the
global aggregation mechanism of graph transformers with the localized
message-passing paradigm of
GNNs~\cite{cai2023connection,southern2025understanding}.
A GVN helps reduce the graph's diameter to two and introduces additional
features that store global information~\cite{southern2025understanding}. Thus,
we can avoid the quadratic complexity of the graph transformers or using
complete graphs while propagating global information through the GVN. Augmenting
a graph with a GVN can be formulated as
\begin{equation}
	\mathrm{VN}(G) = G_{\text{VN}} = (\mathcal{V} \cup \{ v_{N+1} \}, \mathcal{E} \cup \{ (v_j, v_{N+1}) \mid j \in [ N ] \})
\end{equation}
where $v_{N+1}$ denotes the GVN. However, a single virtual node may not
be able to capture all global (or long-range) information.

Sestak\etal
introduced multiple GVNs per graph with learnable features to
attenuate over-squashing for the binding site prediction
task~\cite{sestak2024vnegnn}. Consequently, we can parameterize the global
virtual function $\mathrm{VN}(\cdot)$ as
\begin{equation}
	\begin{split}
		\mathrm{VN}(G; k) = G_{\text{VN}_k} = (\mathcal{V} \cup  \{ v_{N+m} \mid m \in [k] \}, \\
		\mathcal{E} \cup \bigcup_{i=1}^k \{ (v_j, v_{N+i}) \mid j \in [ N ] \})
	\end{split}
\end{equation}
where $k$ is the number of GVNs.

Despite the efficacy of GVNs in creating short circuits for
nodes to receive long-range information while maintaining linear
complexity~\cite{cai2023connection,sestak2024vnegnn}, several limitations
persist that we aim to address in this study. GVNs introduce
significant computational costs when applied to large graphs by adding $k \cdot
	| \mathcal{V} |$ additional edges. Beyond computational concerns, these nodes
inherently modify the global topology, causing deviation from the original graph
structure, though less substantially than graph rewiring techniques. Another
limitation involves the potentially large model width required to store global information effectively~\cite{cai2023connection}.

\begin{figure*}[t!]
	\centering

	\subfloat[Selecting central nodes and adding virtual nodes\label{fig:vn_before}]{%
		\includegraphics[width=0.75\textwidth]{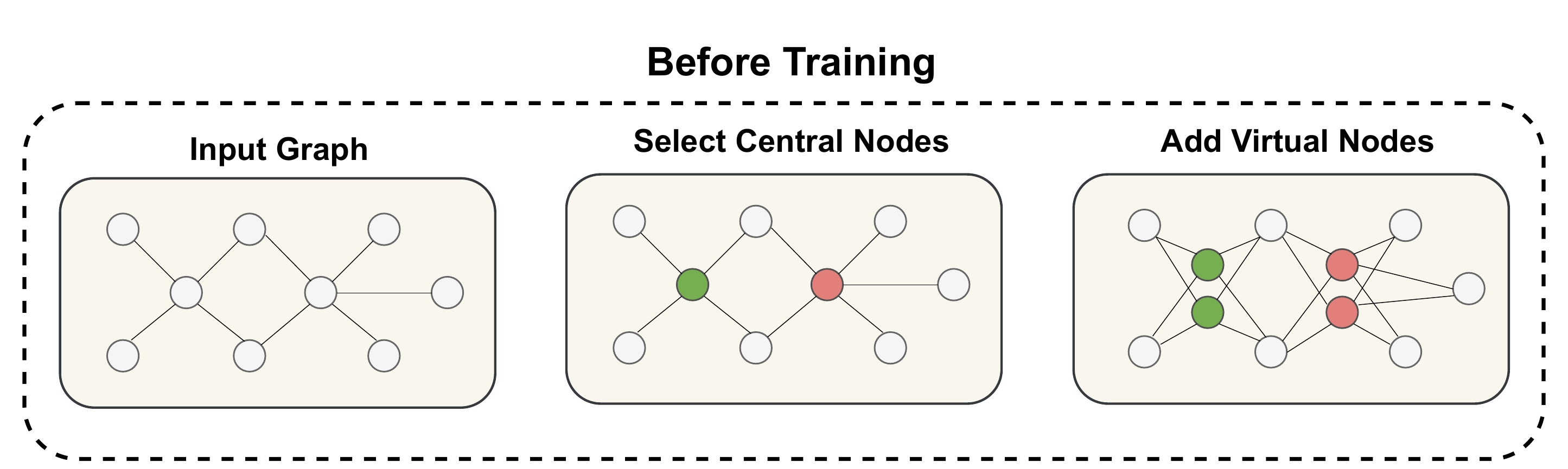}
	}


	\subfloat[Assigning trainable embeddings to virtual nodes\label{fig:vn_during}]{%
		\includegraphics[width=0.75\textwidth]{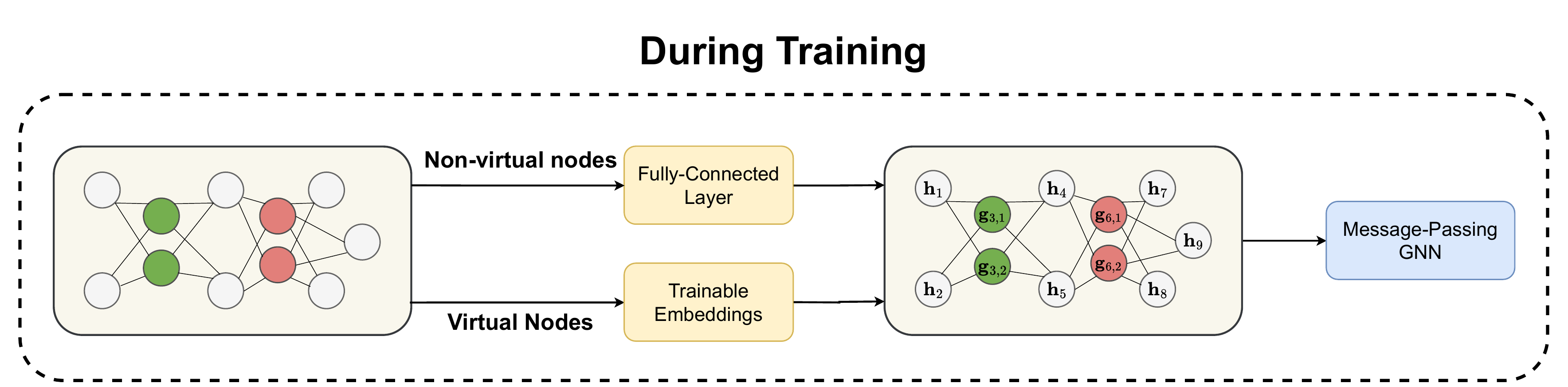}
	}

	\caption{Procedure for adding virtual nodes to a graph. In \ref{fig:vn_before}, we select $n_s$ central nodes and add $n_c$ virtual nodes for each central node, copying the central node's edges to these virtual nodes. \ref{fig:vn_during} shows the process of assigning trainable embeddings to each virtual node before feeding the graph into the GNN.}
	\label{fig:vn_process}
\end{figure*}

\section{Proposed Method: Local Virtual Nodes}
\label{sec:lvn}

This study aims to incorporate local virtual nodes (LVNs) to address bottlenecks without significant changes to the
global graph structure and the elimination of any existing edges. As Figure~\ref{fig:vn_process} shows, the proposed framework
aims to support long-range information flow both before training and during
training. The procedure before training involves obtaining central nodes and
augmenting the input graph with LVNs to introduce more
pathways through central nodes and expand their representation capacity. During
training, trainable LVN embeddings are updated to enable task-specific
information flow between distant nodes. Thus, the proposed framework can be
divided into three essential steps: obtaining central nodes, integrating LVNs,
and message-passing with trainable LVN embeddings. The following sections
present the details of each step.

\subsection{Obtaining Central Nodes}

As discussed earlier, it is costly to identify bottlenecks in a graph. On the other hand, the correlation between the node centrality and bottlenecks in the graphs has been discussed in the literature~\cite{topping2022understanding, choi2024panda}. Therefore, we also consider the centrality measure to obtain regions with a high risk of bottlenecks. A subset of nodes representing common stations connecting most nodes is obtained by leveraging the centrality metrics following the footsteps of Choi {\it et al.} in the rewiring study coined PANDA~\cite{choi2024panda}. However, unlike PANDA, centrality metrics with linear complexity in terms of the number of nodes and edges are preferred in this study.

We utilize a centrality function, $C: \mathcal{V} \rightarrow \mathbb{R} $, to obtain the centrality scores of nodes. Three centrality functions based on degree, PageRank~\cite{page1999pagerank}, and label
propagation~\cite{raghavan2007near} are considered. Degree centrality and PageRank centrality functions are implemented as in PANDA~\cite{choi2024panda}.
Alongside these standard centrality measures, we also implement a centrality
function that applies label propagation to cluster the graph into communities
and calculates the out-community degree of each node. Then, for each community,
we select the node with the highest out-community degree as the central node.
The rationale behind this criterion is that a node with the maximum connections to external communities is more likely to be a source of a bottleneck. The centrality scores are then ranked, and the top $n_s$ nodes with the highest scores are identified. The set of chosen most central nodes is denoted by $\mathcal{C}$.

\subsection{Local Virtual Node Integration}

In contrast to the studies treating virtual nodes as entirely new nodes~\cite{southern2025understanding,zhang2024improving,hwang2022an,cai2023connection,sestak2024vnegnn}, we characterize LVNs as support nodes that introduce new pathways through central regions and increase the feature capacity of existing nodes. Thus, we augment each central node in $\mathcal{C}$ with a set of LVNs, which we call an LVN group. Thus, an LVN group represents the original central node. The size of the LVN group, $n_c$, is designed to be a hyperparameter. This hyperparameter may be tuned depending on the graph topology in the dataset. The nodes in a severe bottleneck region may benefit from a higher number of LVNs. On the other hand, only two LVNs may also be sufficient for central nodes in a well-connected community. Each LVN in the same group operates independently of each other and
receives information from neighbors of the original node while also propagating
features.

\begin{figure}
	\centering
	\includegraphics[width=0.85\columnwidth]{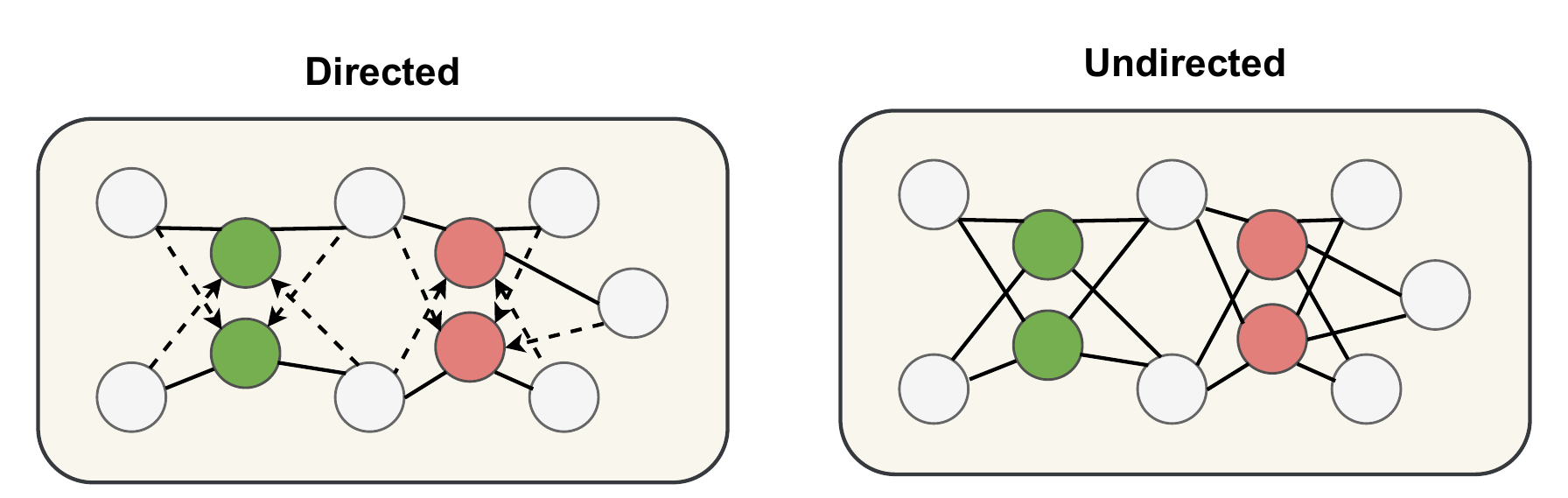}
	\caption{A graph augmented with LVNs directed (Left) and undirected (Right) edge distribution methods. Dotted lines represent directed edges.}
	\label{fig:vn_directed}
\end{figure}

The LVN group of a central node is integrated by distributing the edges of the original node among the LVNs. Several strategies could be employed for this purpose. We propose the standard strategy, a straightforward approach where each LVN is connected to the neighbors of its central node. The standard strategy is illustrated by the Before Training step in Figure~\ref{fig:vn_process}.
However, this standard strategy might have two immediate drawbacks. First, the number of edges might significantly increase when the number of LVNs grows. Secondly, over-smoothing might occur since the central node neighbors receive information from all the LVNs in a group. Therefore, we propose an additional approach, a directed edge distribution strategy illustrated in Figure~\ref{fig:vn_directed}, where each neighbor of the original central node connects to only one LVN from the group. Aside from alleviating over-smoothing, the directed approach enables virtual nodes to specialize by
sending features to only a subset of neighbors. However, in the directed approach, each LVN still receives information from all the neighbors of the corresponding central node, which is consistent with the standard edge distribution strategy. This asymmetry creates a
directed alternative to the standard approach.

Analogous to GVNs, we can formalize the proposed LVN framework. Suppose $c(\cdot)$ assigns an index to each central node in
$\mathcal{C}$ from the set $[n_s]$. The LVN integration with the
given hyperparameters can be expressed as follows
\begin{equation}
	\mathrm{LVN}(G; \mathcal{C}, n_c) = G_{\text{LVN}} = (\mathcal{V}_{\text{LVN}}, \mathcal{E}_{\text{LVN}})
\end{equation}
where $\mathcal{V}_{\text{LVN}}$ and $\mathcal{E}_{\text{LVN}}$ indicate the augmented
set of nodes and edges, respectively.

The augmented set of nodes, $\mathcal{V}_{\text{LVN}}$, is defined as
\begin{equation}
	\mathcal{V}_{\text{LVN}} = \mathcal{V} \cup \bigcup_{j \in \mathcal{C}} \{ v_{N+c(j)m} \mid m \in [n_c]  \}
\end{equation}
where $\mathcal{V}$ is augmented with $n_s \cdot n_c$ nodes. Whereas, the
updated edge set $\mathcal{E}_{\text{LVN}}$ can be defined as
\begin{equation}
	\begin{split}
		\mathcal{E}_{\text{LVN}} & = \mathcal{E} \cup \bigcup_{\substack{i \in \mathcal{C}, \\ m \in [n_c]}} \{ (j, v_{N+c(i)m}) \mid j \in \mathcal{N}(i) \} \\
		                         & \quad \cup \bigcup_{\substack{i,j \in \mathcal{C},       \\ (i, j) \in \mathcal{E}}} \{ (v_{N+c(i)y}, v_{N+c(j)z}) \mid y,z \in [n_c] \}
	\end{split}
\end{equation}
where we merge the existing edge set with the LVN edges originating
from central nodes. The first union operator corresponds to assigning the
original neighborhood as LVN neighbors. The second union operator
represents edges from the corner case where the selected central
nodes are adjacent. The same formalization applies to the directed
edge distribution strategy. Finally, as we introduce the LVN groups as a replacement for the
central nodes, we remove them from the graph by taking the induced subgraph of
$G_{\text{LVN}}$ for the set $\mathcal{V} \setminus \mathcal{C}$. More formally,
we feed the graph, $G_{\text{LVN}}[\mathcal{V} \setminus \mathcal{C}]$, to the
GNN.

Selecting central nodes and LVN integration constitute the steps applied before training in our proposed framework. This step aims to ensure that the new local pathways are created around bottleneck regions determined through centrality, while the global graph topology driven by the domain knowledge is not significantly distorted. The next step involves populating the LVNs with embeddings and training a GNN architecture with the LVN augmented input graph to tackle graph-level and node-level tasks.

\subsection{Message-Passing with Trainable LVN Embeddings}

We hypothesize that each LVN in the same group should have distinct features to expand the information propagation capacity. Copying the original input features of the central node to its LVNs would not facilitate diversity among the updated LVN node representations during message-passing. To encourage learning diverse LVN representation in a group, we
propose to assign trainable feature vectors to each LVN, denoted by $\mathbf{P}^{n_c \times D}$, where $D$ represents the hidden layer
dimensionality of the GNN.

The trainable LVN embeddings are shared across each LVN group to avoid increasing the number of parameters if we select more central nodes. More importantly, sharing LVN embeddings across groups provides communication between distant nodes. Since each group is associated with a different central region on the graph, shared LVN embeddings enable knowledge sharing between regions, which is essential for long-range propagation. During training, trainable embeddings are updated using multiple
feedback received from possibly non-contiguous regions of the graph. Therefore,
the trainable embeddings in an LVN group contain long-range information
independent of the number of message-passing layers. Thus, the shared trainable LVN embeddings allow us to extend the receptive field of GNNs without necessarily adding more message-passing layers.

Let us formally define how we construct the initial set of node features,
$\mathbf{x}^{(0)}$. We feed the input node features to a fully-connected layer
to map them to a $D$-dimensional latent space. Then each LVN is assigned to one
of the trainable embeddings. Each LVN in a virtual node group receives an index
from $[n_c]$, indicating its the position within the virtual node group. We
experiment with two approaches to utilizing shared embeddings. One is named ``replace" mode, in which we only assign the LVN embeddings as initial features
of the virtual node. This approach works well in datasets with no input features
and even in some settings where input features are present, hinting that the
virtual node embeddings and neighbor features combined have enough quality to
represent the central node. The second approach is the ``addition" mode that
sets the initial virtual node features as the summation of the original features
and the virtual node embeddings, much like the positional encodings in
Transformers~\cite{vaswani2017attention}. The initial representation of a
non-virtual node is determined as
\begin{equation}
	\mathbf{h}_v = \mathbf{W}_{\text{in}}^{T} \mathbf{x}_v + \mathbf{b}_{\text{in}}
\end{equation}
where $\mathbf{W}_{\text{in}} \in \mathbb{R}^{F \times D}$,
$\mathbf{b}_{\text{in}} \in \mathbb{R}^{D}$ are the weights and biases of the
fully-connected layer. As for virtual nodes, we set the initial representation
of the $i$-th virtual node in the LVN group representing the central node $v$ as
\begin{equation}
	\mathbf{g}_{v,i} =
	\begin{cases}
		\mathbf{p}_{i},                                                                    & \text{replace mode} \\
		\mathbf{W}_{\text{in}}^{T} \mathbf{x}_v + \mathbf{b}_{\text{in}} + \mathbf{p}_{i}, & \text{add mode}
	\end{cases}
\end{equation}
where $\mathbf{p}_i$ denotes the $i$-th row of the embedding matrix. Depending
on whether the node is a virtual node or not, we feed either $\mathbf{g}_{v,i}$
or $\mathbf{h}_v$ as the input features $\mathbf{x}_v^{(0)}$ to the first
message-passing layer.

\section{Experiments}
\label{chap:experiments}

This section presents the performance of the proposed LVN framework on graph classification and node classification
benchmarks. Additionally, we report metrics demonstrating improved graph
connectivity and the effectiveness of virtual node embeddings.

\subsection{Datasets}

To evaluate the proposed method on graph-level tasks, we used the following
graph classification datasets from the TUDataset~\cite{morris2020tudataset}:
REDDIT-BINARY, IMDB-BINARY, MUTAG, PROTEINS, ENZYMES, and COLLAB. REDDIT-BINARY,
IMDB-BINARY, and COLLAB contain social networks. ENZYMES and PROTEINS are
bioinformatics datasets of macromolecules represented as
graphs~\cite{morris2020tudataset}. MUTAG is a dataset of small
molecules~\cite{morris2020tudataset}. MUTAG, ENZYMES, and PROTEINS have one-hot
feature vectors,  whereas REDDIT-BINARY, IMDB-BINARY, and COLLAB do not have node
features. For these datasets, we add a single constant feature with a value of
one, following the experimental setup of FoSR and
PANDA~\cite{karhadkar2023fosr,choi2024panda}. The statistics of the datasets are
provided in Table~\ref{tab:dataset_stats_graph}. We also used six single-graph datasets designed for node classification tasks.
Chameleon, Cornell, Wisconsin, and Texas are web networks comprising webpages as
nodes and their incoming and outgoing hyperlinks as edges~\cite{pei2020geomgcn}.
Node features are bag-of-words representations of webpage
contents~\cite{pei2020geomgcn}. In addition, we test our approach on Cora and
Citeseer, citation graphs where nodes represent research papers and edges
represent citations between papers~\cite{sen2008collective}. As with the web
networks, node features are the bag-of-words representation of paper
abstracts~\cite{sen2008collective}.

\begin{table}[t!]
	\centering
	\caption{Statistics of the graph classification benchmark datasets.}
	\resizebox{\linewidth}{!}{
		\setlength{\tabcolsep}{1.5pt}
		\begin{tabular}{lcccc}
			\toprule
			\textbf{Dataset} & \textbf{\# Graphs} & \textbf{Avg. Degree} & \textbf{Avg. \# Nodes} & \textbf{Avg. \# Edges} \\
			\midrule
			REDDIT-BINARY    & $2{,}000$          & $2.34$               & $429.63$               & $995.51$               \\
			IMDB-BINARY      & $1{,}000$          & $8.89$               & $19.77$                & $193.06$               \\
			MUTAG            & $188$              & $2.19$               & $17.93$                & $39.59$                \\
			ENZYMES          & $600$              & $3.86$               & $32.63$                & $124.27$               \\
			PROTEINS         & $1{,}113$          & $3.73$               & $39.06$                & $145.63$               \\
			COLLAB           & $5{,}000$          & $37.37$              & $74.49$                & $4{,}914.43$           \\
			\bottomrule
		\end{tabular}
	}
	\label{tab:dataset_stats_graph}
\end{table}
\begin{table}[t!]
	\centering
	\caption{Statistics of the node classification benchmark datasets.}
	\resizebox{0.8\linewidth}{!}{
		\begin{tabular}{lccc}
			\toprule
			\textbf{Dataset} & \textbf{\# Nodes} & \textbf{\# Edges} & \textbf{\# Features} \\
			\midrule
			Chameleon        & $2{,}277$         & $36{,}051$        & $2{,}325$            \\
			Cornell          & $183$             & $295$             & $1{,}703$            \\
			Texas            & $183$             & $309$             & $1{,}703$            \\
			Wisconsin        & $251$             & $499$             & $1{,}703$            \\
			Cora             & $2{,}708$         & $10{,}556$        & $1{,}433$            \\
			Citeseer         & $3{,}327$         & $9{,}104$         & $3{,}703$            \\
			\bottomrule
		\end{tabular}
	}
	\label{tab:dataset_stats_node}
\end{table}

\subsection{Baselines}

We compare our approach to state-of-the-art graph rewiring and related methods
in the literature. DIGL~\cite{gasteiger2019diffusion} represents one of the
first rewiring approaches, learning a graph diffusion matrix by constructing a
weighted sum of exponents of the symmetrically normalized adjacency matrix used
in GCN~\cite{kipf2017semisupervised}. The resulting matrix represents a locally
rewired graph and is used for message-passing rather than the original adjacency
matrix~\cite{gasteiger2019diffusion}. On the other hand, recent graph-rewiring approaches focus on optimizing metrics that measure over-squashing, such as the spectral gap. FoSR~\cite{karhadkar2023fosr} is an iterative
algorithm that aims to maximize the spectral gap by adding edges that increase
it.

Unlike global metrics like spectral gap, graph curvature is also used to identify bottlenecks. SDRF~\cite{topping2022understanding} is the first study that performs graph rewiring
by iteratively adding edges that improve the curvature of the lowest-curvature
edge, likely to contain bottlenecks, and remove high-curvature edges to avoid redundancy.
BORF~\cite{nguyen2023revisiting} uses an improved curvature measure,
Ollivier-Ricci curvature, to rewire graphs while avoiding both over-smoothing
and over-squashing~\cite{nguyen2023revisiting}. LASER~\cite{barbero2024localityaware} aims to reach the performance of spectral
rewiring approaches while preserving the graph's locality and sparsity. In
addition to preserving locality, LASER applies sequential rewiring, resulting in
multiple snapshots of the graph used during the forward pass. The authors assigned
each snapshot a separate GCN weight at the last layer; therefore, LASER uses
more GCN layers than the other baselines. We use the LASER's setup to ensure
a fair comparison. We also included the recently proposed width expansion method
PANDA~\cite{choi2024panda} that assigns a higher width to central nodes and
applies different message-passing operations based on the type of source and
destination nodes (low-width or high-width). Finally, following Alon and Yahav~\cite{alon2021on}, we also implement an
approach using a complete graph at the last layer of the GNN, which we refer to
as Last Layer FA. Each baseline is trained from scratch using the reported hyperparameters by the authors.

\subsection{Learning Setup}

To ensure fair comparisons across baselines, we maintained a consistent architecture
and optimization hyperparameters within each task while leveraging official
implementations from
BORF\footnote{\url{https://github.com/hieubkvn123/revisiting-gnn-curvature}},
LASER\footnote{\url{https://github.com/Fedzbar/laser-release}}, and
PANDA\footnote{\url{https://github.com/jeongwhanchoi/PANDA}} repositories. GCN~\cite{kipf2017semisupervised} was used in all the experiments as the backbone architecture with ReLU activation functions and dropout
layers (rate $0.5$) between consecutive layers. We employed the Adam optimizer~\cite{kingma2015adam} with
learning rate $10^{-3}$ and early stopping based on validation accuracy with a
patience of 100 epochs.

\subsubsection{Graph Classification}
Following the setup of FoSR~\cite{karhadkar2023fosr} for graph classification,
we trained each baseline with a GCN containing four hidden layers and hidden
dimensionality of $64$. One exception is LASER~\cite{barbero2024localityaware},
which by design contains more GCN layers than other baselines, and also uses
batch normalization~\cite{ioffe2015batch} and removes dropout. We randomly split the
graphs in the datasets into 80\% training, 10\% validation, and 10\% test. We used
the best rewiring hyperparameters for DIGL, SDRF, and BORF reported by
Nguyen\etal\cite{nguyen2023revisiting}.

\subsubsection{Node Classification}
In node classification experiments, we used a GCN with three hidden layers and
hidden dimensionality of $128$, the same setup as BORF~\cite{nguyen2023revisiting}.
We split the nodes in each graph into 60\% training, 20\% validation, and 20\%
test sets at random. The LASER results are unavailable for this task as the
original study focused exclusively on graph classification and did not provide
node classification implementations or hyperparameters. We again used the
rewiring hyperparameters reported by Nguyen\etal for DIGL, SDRF, and
BORF~\cite{nguyen2023revisiting}. Finally, we trained PANDA with the reported
hyperparameters~\cite{choi2024panda}, but for the unreported dataset Chameleon,
we set top\_k to 50 (same as other node classification datasets) and search for
best centrality and $p_{\text{high}}$ in the following spaces: $ \{
	\text{betweenness}, \text{degree} \}$, $\{ 160, 192\}$. Note that, unlike other
baselines, Choi\etal uses a hidden dimension of $64$; therefore, we followed their
setup for retraining PANDA. In our approach, as we replace central nodes with LVN groups, we apply mean pooling across each LVN group to obtain a single prediction for the original node.

\subsection{Predictive Performance}

\begin{table*}[t!]
	\centering
	\vskip\baselineskip \caption{Graph classification performance of LVN compared with baselines. All methods are trained with a GCN backbone. OOM indicates Out of
		Memory. Colors represent the \textcolor{first}{First},
		\textcolor{second}{Second} and \textcolor{third}{Third} best performing method.}
	\resizebox{\linewidth}{!}{
		\begin{tabular}{lccccccc}
			\toprule
			\textbf{Model}                        & \textbf{REDDIT-BINARY}                          & \textbf{IMDB-BINARY}                            & \textbf{MUTAG}                                  & \textbf{ENZYMES}                                & \textbf{PROTEINS}                               & \textbf{COLLAB}                                 & \textbf{Avg. Rank} \\
			\midrule
			GCN~\cite{kipf2017semisupervised}     & $56.530 \pm 2.603$                              & $50.020 \pm 1.309$                              & $71.400 \pm 3.125$                              & $28.033 \pm 2.019$                              & $71.410 \pm 1.077$                              & $48.556 \pm 2.339$                              & $7.17$             \\
			Last Layer FA~\cite{alon2021on}       & $58.680 \pm 2.421$                              & $49.900 \pm 1.574$                              & $69.400 \pm 3.245$                              & $27.267 \pm 1.810$                              & $71.964 \pm 1.362$                              & $51.140 \pm 1.107$                              & $7.17$             \\
			DIGL~\cite{gasteiger2019diffusion}    & $50.440 \pm 0.938$                              & $50.280 \pm 1.261$                              & $76.000 \pm 3.072$                              & $27.467 \pm 1.682$                              & $70.929 \pm 1.290$                              & $32.920 \pm 5.231$                              & $7.50$             \\
			SDRF~\cite{topping2022understanding}  & $62.720 \pm 2.701$                              & $49.760 \pm 1.338$                              & $73.300 \pm 2.617$                              & $28.533 \pm 1.616$                              & $71.054 \pm 0.925$                              & $46.548 \pm 2.576$                              & $7.17$             \\
			FoSR~\cite{karhadkar2023fosr}         & $70.020 \pm 0.911$                              & $50.120 \pm 1.263$                              & $80.400 \pm 2.430$                              & $23.567 \pm 1.313$                              & $72.071 \pm 1.163$                              & $48.772 \pm 2.308$                              & $5.83$             \\
			BORF~\cite{nguyen2023revisiting}      & OOM                                             & $47.280 \pm 1.473$                              & $76.500 \pm 3.397$                              & $25.767 \pm 1.711$                              & $70.339 \pm 0.918$                              & OOM                                             & $8.75$             \\
			PANDA~\cite{choi2024panda}            & $80.080 \pm 1.241$                              & \textcolor{third}{\textbf{$64.080 \pm 1.319$}}  & \textcolor{second}{\textbf{$83.000 \pm 1.876$}} & \textcolor{second}{\textbf{$32.067 \pm 1.856$}} & \textcolor{first}{\textbf{$75.750 \pm 0.944$}}  & \textcolor{third}{\textbf{$68.060 \pm 0.616$}}  & $2.50$             \\
			LASER~\cite{barbero2024localityaware} & \textcolor{first}{\textbf{$88.130 \pm 0.547$}}  & $57.520 \pm 1.438$                              & $76.842 \pm 2.459$                              & \textcolor{first}{\textbf{$33.800 \pm 1.758$}}  & $70.625 \pm 1.264$                              & \textcolor{first}{\textbf{$71.880 \pm 0.567$}}  & $3.50$             \\
			\textbf{LVN (Undirected)}             & \textcolor{third}{\textbf{$83.440 \pm 0.775$}}  & \textcolor{second}{\textbf{$66.620 \pm 1.513$}} & \textcolor{third}{\textbf{$82.333 \pm 2.149$}}  & $31.367 \pm 1.376$                              & \textcolor{third}{\textbf{$73.189 \pm 0.882$}}  & \textcolor{second}{\textbf{$71.520 \pm 0.661$}} & $2.83$             \\
			\textbf{LVN (Directed)}               & \textcolor{second}{\textbf{$85.620 \pm 0.990$}} & \textcolor{first}{\textbf{$68.720 \pm 1.477$}}  & \textcolor{first}{\textbf{$84.778 \pm 2.902$}}  & \textcolor{third}{\textbf{$31.400 \pm 2.095$}}  & \textcolor{second}{\textbf{$74.360 \pm 1.105$}} & $65.588 \pm 1.064$                              & $2.17$             \\
			\bottomrule
		\end{tabular}
	}
	\label{tab:benchmark_graph}
\end{table*}
\begin{table*}[t!]
	\centering
	\vskip\baselineskip \caption{Node classification performance of LVN compared with baselines. All methods are trained with a GCN backbone. Colors represent the
		\textcolor{first}{First}, \textcolor{second}{Second} and
		\textcolor{third}{Third} best performing method.} 
	\resizebox{\linewidth}{!}{
		\begin{tabular}{lccccccc}
			\toprule
			\textbf{Model}                       & \textbf{CORA}                                 & \textbf{CITESEER}                             & \textbf{TEXAS}                                & \textbf{CORNELL}                              & \textbf{WISCONSIN}                            & \textbf{CHAMELEON}                            & \textbf{Avg. Rank} \\
			\midrule
			GCN~\cite{kipf2017semisupervised}    & \textcolor{second}{\textbf{$87.59 \pm 0.22$}} & $73.58 \pm 0.27$                              & $52.00 \pm 2.10$                              & $48.27 \pm 1.82$                              & $50.00 \pm 1.28$                              & $61.12 \pm 0.46$                              & $5.00$             \\
			DIGL~\cite{gasteiger2019diffusion}   & \textcolor{first}{\textbf{$87.61 \pm 0.29$}}  & $73.76 \pm 0.35$                              & $51.57 \pm 1.78$                              & \textcolor{third}{\textbf{$48.65 \pm 1.50$}}  & \textcolor{second}{\textbf{$51.33 \pm 1.23$}} & \textcolor{first}{\textbf{$62.08 \pm 0.54$}}  & $3.00$             \\
			SDRF~\cite{topping2022understanding} & $87.51 \pm 0.29$                              & $73.74 \pm 0.34$                              & $51.51 \pm 1.71$                              & $47.62 \pm 1.75$                              & \textcolor{third}{\textbf{$51.14 \pm 1.36$}}  & $61.19 \pm 0.55$                              & $4.83$             \\
			FoSR~\cite{karhadkar2023fosr}        & $86.88 \pm 0.31$                              & $73.88 \pm 0.32$                              & $52.86 \pm 1.72$                              & $44.16 \pm 1.46$                              & $51.10 \pm 1.42$                              & \textcolor{third}{\textbf{$61.23 \pm 0.53$}}  & $4.50$             \\
			BORF~\cite{nguyen2023revisiting}     & \textcolor{third}{\textbf{$87.58 \pm 0.33$}}  & $73.32 \pm 0.28$                              & $49.95 \pm 1.43$                              & \textcolor{first}{\textbf{$50.65 \pm 1.62$}}  & $50.86 \pm 1.23$                              & \textcolor{second}{\textbf{$61.49 \pm 0.53$}} & $4.50$             \\
			PANDA~\cite{choi2024panda}           & $86.69 \pm 0.43$                              & \textcolor{second}{\textbf{$74.97 \pm 0.48$}} & \textcolor{third}{\textbf{$54.11 \pm 2.15$}}  & $42.49 \pm 2.59$                              & $50.47 \pm 2.04$                              & $50.55 \pm 0.70$                              & $5.83$             \\
			\textbf{LVN (Undirected)}            & $86.73 \pm 0.40$                              & \textcolor{first}{\textbf{$74.98 \pm 0.48$}}  & \textcolor{first}{\textbf{$61.37 \pm 2.03$}}  & \textcolor{second}{\textbf{$49.21 \pm 1.89$}} & \textcolor{first}{\textbf{$55.26 \pm 1.99$}}  & $60.09 \pm 0.71$                              & $3.17$             \\
			\textbf{LVN (Directed)}              & $86.90 \pm 0.38$                              & \textcolor{third}{\textbf{ $74.30 \pm 0.51$}} & \textcolor{second}{\textbf{$56.84 \pm 2.16$}} & $42.90 \pm 1.90$                              & $49.10 \pm 1.55$                              & $60.43 \pm 0.67$                              & $5.17$             \\
			\bottomrule
		\end{tabular}
	}
	\label{tab:node_classification}
\end{table*}

Graph classification and node classification  performances of baselines that
combat over-squashing are given in Table~\ref{tab:benchmark_graph} and
Table~\ref{tab:node_classification}, respectively. All results report mean test accuracy with $95\%$ confidence intervals across 50 random splits.

Table~\ref{tab:benchmark_graph} shows that the proposed framework outperforms
established graph rewiring methods designed to address bottlenecks and the
state-of-the-art width-expansion approach PANDA by a large margin. In
particular, the LVNs excel on social network datasets such as REDDIT-BINARY,
IMDB-BINARY and COLLAB, which feature central nodes connecting communities that
can create bottleneck problems. Additionally, the trainable LVN embeddings are
particularly important for learning distinct features on social graph datasets,
which lack input node features. We also substantially outperform most of the
graph rewiring methods in bioinformatics datasets like MUTAG, ENZYMES, and
PROTEINS. While our approach shows competitive performance on node
classification, the improvements are less salient than in graph classification.
We posit that the node classification problem is primarily local and suffers
more from over-smoothing than over-squashing.
Table~\ref{tab:node_classification} presents the node classification performance
where we can see that none of the rewiring algorithms achieve substantial
improvements over standard GCN.

\subsection{Effective Resistance}

The proposed LVN approach is also evaluated using \textit{effective resistance}, an analogy from electrical networks, which quantifies the connectivity between two nodes in a graph~\cite{black2023understanding}. Effective resistance has been leveraged in the graph rewiring literature~\cite{banerjee2022oversquashing,nguyen2023revisiting}. Furthermore, Black\etal theoretically proved that effective resistance is
related to over-squashing and proposed a rewiring algorithm that reduces total effective resistance~\cite{black2023understanding}.

Effective resistance between a pair of nodes $u$ and $v$ is formally defined as
\begin{equation}
	\mathcal{R}_{u,v} = (\mathbf{1}_u-\mathbf{1}_v)^{T} \mathbf{L}^{+} (\mathbf{1}_u-\mathbf{1}_v)
\end{equation}
where $\mathbf{1}_u$ denotes a vector with entries all zero but one at $u$ and
$\mathbf{L}^{+}$ is the Moore-Penrose pseudoinverse of the Laplacian
matrix~\cite{banerjee2022oversquashing,black2023understanding}. Then the total
effective resistance of a graph is defined as
\begin{equation}
	\mathcal{R}_{\text{tot}}(G) = \frac{1}{2}\sum_{u,v \in \mathcal{V}} \mathcal{R}_{u,v} = N \sum_{i=2}^{N} \frac{1}{\lambda_i}
\end{equation}
where $\lambda_i$ denotes the $i$-th eigenvalue of the Laplacian
$\mathbf{L}$~\cite{ghosh2008minimizing}. Black\etal computed the total effective
resistance as new edges are added to the graph~\cite{black2023understanding}.
Since $\mathcal{R}_{\text{tot}}(G)$ increases with the number of nodes, directly comparing total effective resistance as LVNs are added would be unfair.
Therefore, we compute a modified version using a fixed subset of nodes. To
ensure a fair comparison, we determine a subset of nodes $S$ that includes the
non-virtual nodes when the number of selected nodes to add virtual nodes is
maximum. Particularly, we independently transform the graph to include virtual
nodes for the following set of $n_s$ values: $\{ 1, 2, 3, 5, 7, 10, 12, 15 \}$.
Let $\mathcal{C}$ denote the set of $15$ most central nodes. For fair comparison
across all cases, we define $\mathcal{S} = \mathcal{V} \setminus \mathcal{C}$ and compute
total effective resistance only between pairs of nodes in $S$.

\begin{figure*}[t!]
	\centering
	\begin{tabular}{ccc}
		\includegraphics[width=0.29\textwidth]{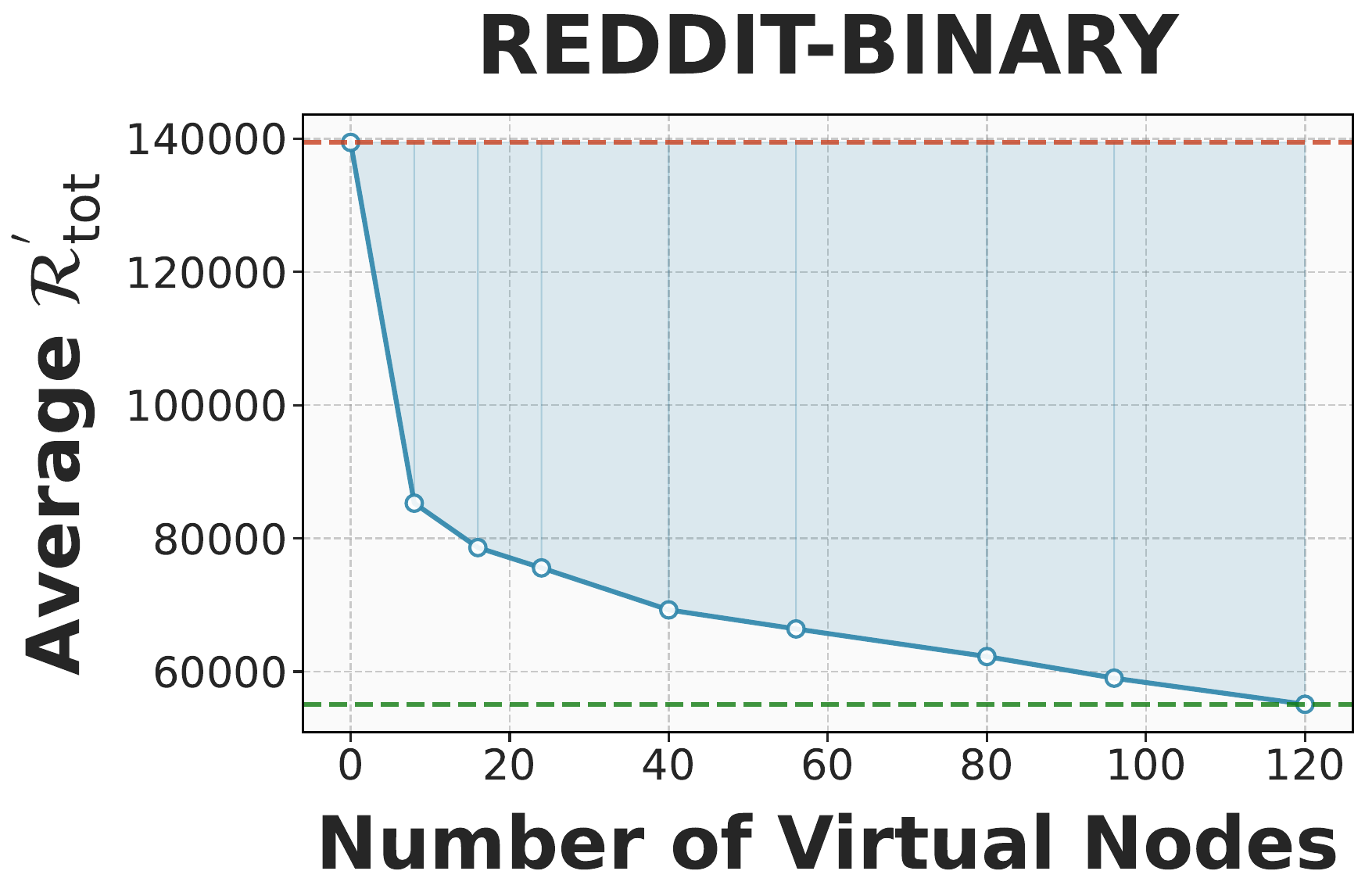} &
		\includegraphics[width=0.29\textwidth]{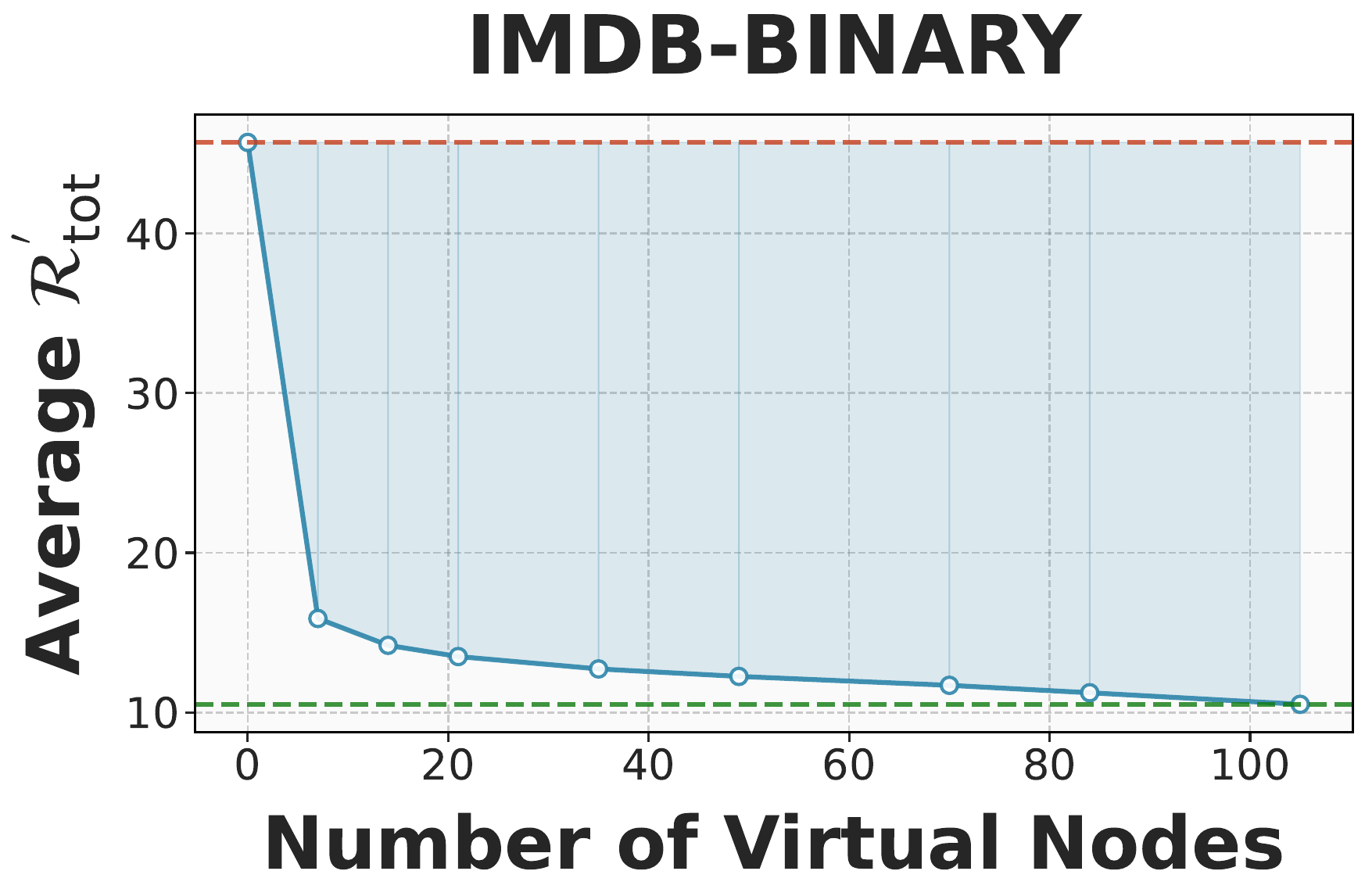}   &

		\includegraphics[width=0.29\textwidth]{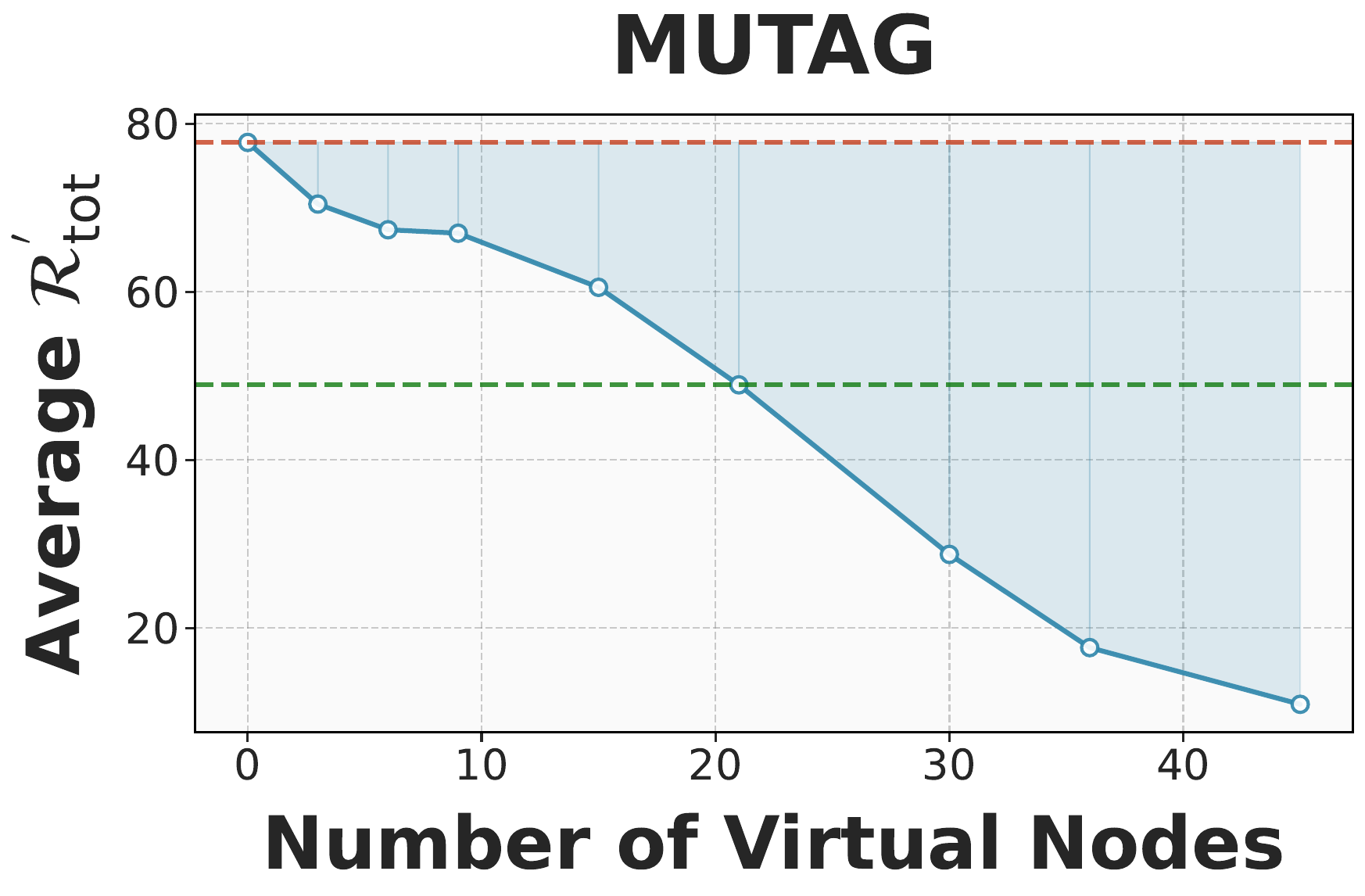}           \\
		\includegraphics[width=0.29\textwidth]{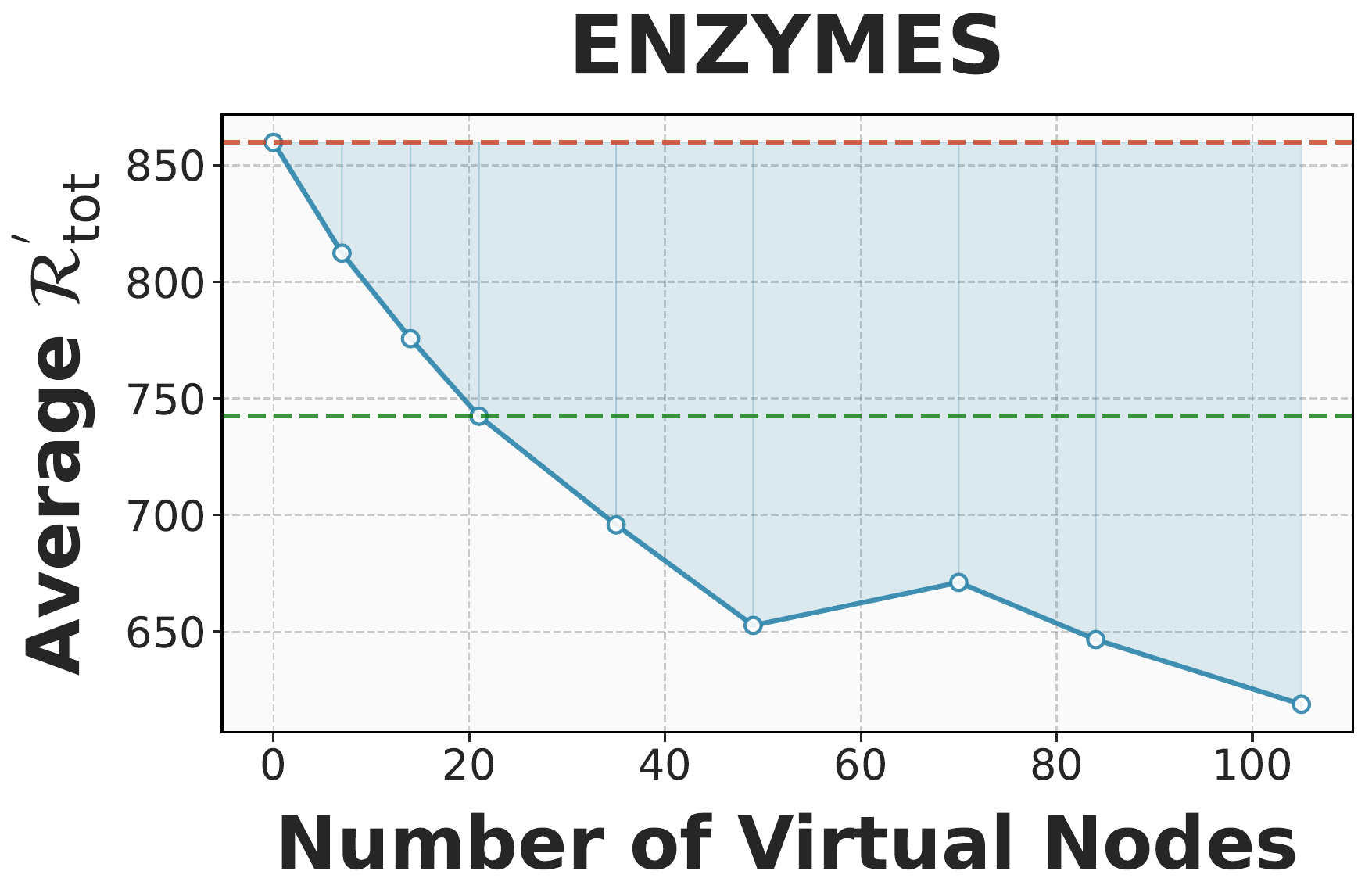}       &

		\includegraphics[width=0.29\textwidth]{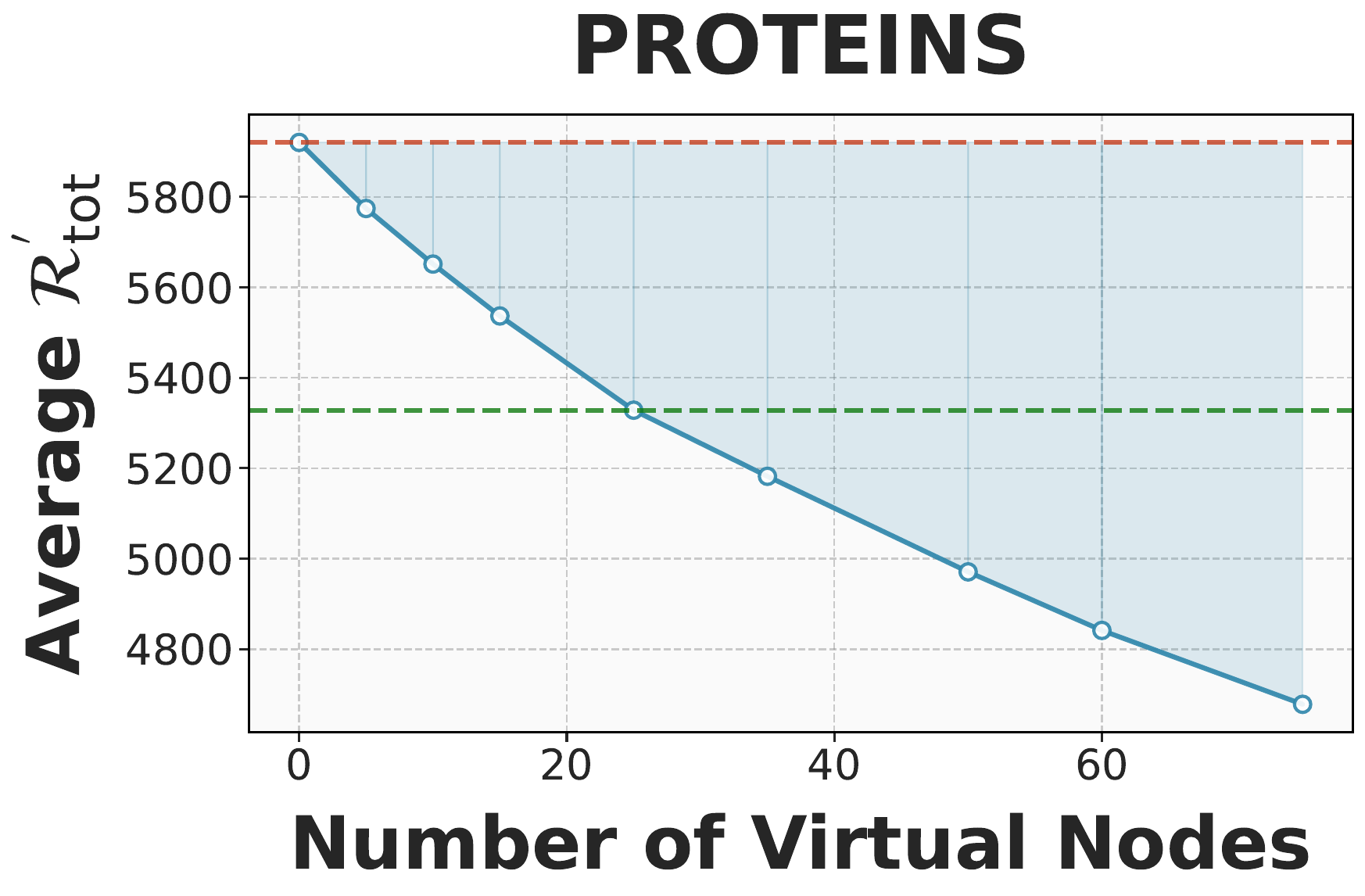}      &
		\includegraphics[width=0.29\textwidth]{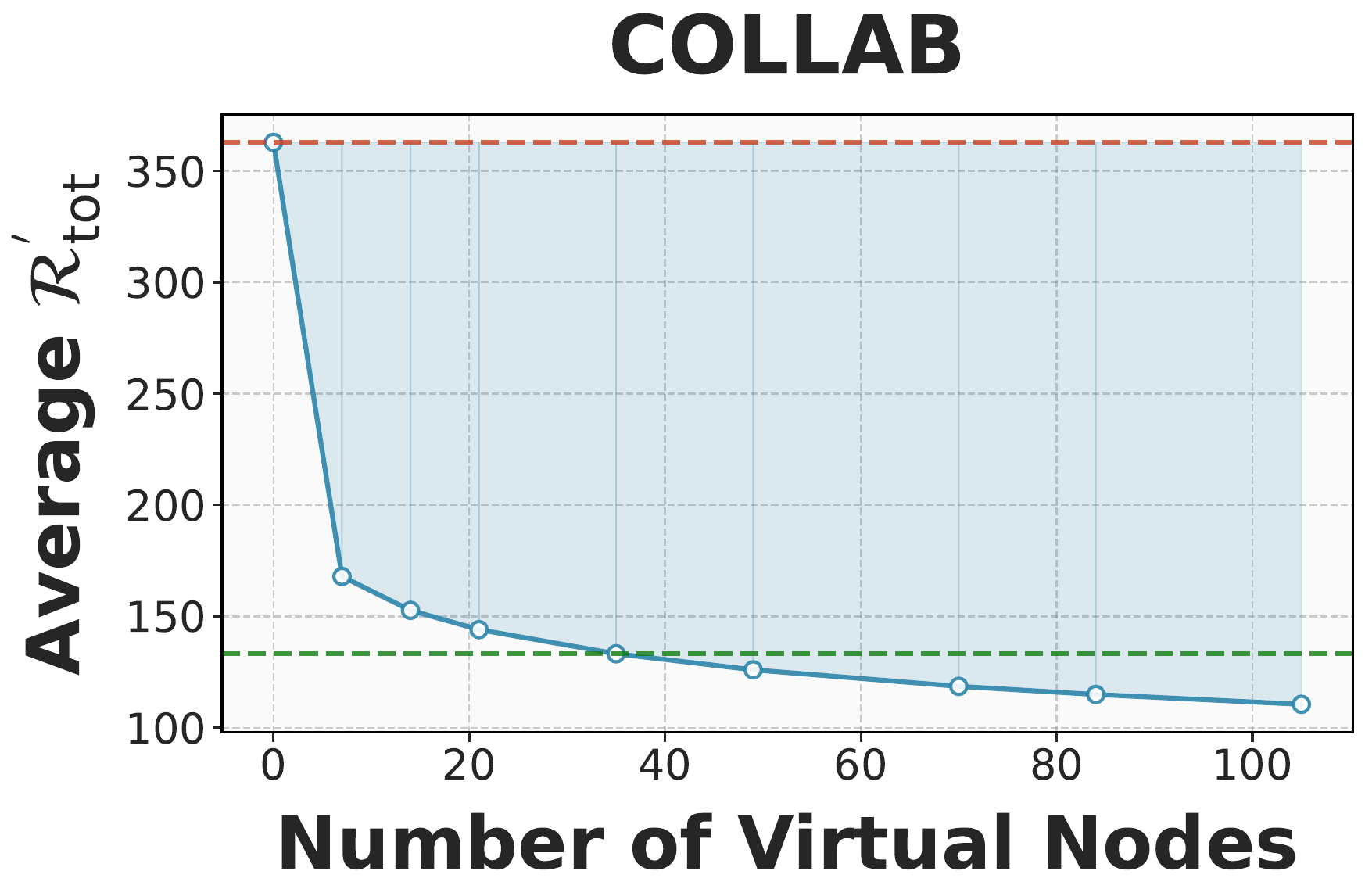}          \\
	\end{tabular}
	\caption{Change in the total effective resistance of the non-virtual node
		set for TUDataset~\cite{morris2020tudataset}.
		\textsf{\textcolor{resblue}{\textemdash$\circ$\textemdash Transformed
				Graph}, \textcolor{resred}{- - Raw Graph}, \textcolor{resgreen}{- - Best
				Configuration}}. Shaded areas show improvement relative to baseline.}
	\label{fig:total_eff_res}
\end{figure*}

Figure~\ref{fig:total_eff_res} shows that adding LVNs improves the connectivity between non-central nodes by reducing the effective resistance. A substantial reduction in the total effective resistance can be observed with
only a few virtual nodes for social network datasets since they usually contain a small subset of central nodes connecting several communities. On the other hand,  a monotonic decrease in total effective resistance is observed for bioinformatics and small molecule datasets with path-like graphs and uniform degree distribution.

\subsection{Change in the Number of Paths}

To further investigate whether adding LVNs improves connectivity, we compared the difference in the number of paths between the original graph and the
one augmented with the LVNs. We computed the number of walks of length exactly $r$ between each node pair, denoted as $\mathbf{A}^r$.
Similar to effective resistance, we computed $\mathbf{A}^r_{ij}$ for each $(i,
	j)$-pair in the non-central node set $\mathcal{S}$ since we aim to see any increase in the number of paths through central nodes.

\begin{figure}[t!]
	\centering
	\includegraphics[width=0.85\columnwidth]{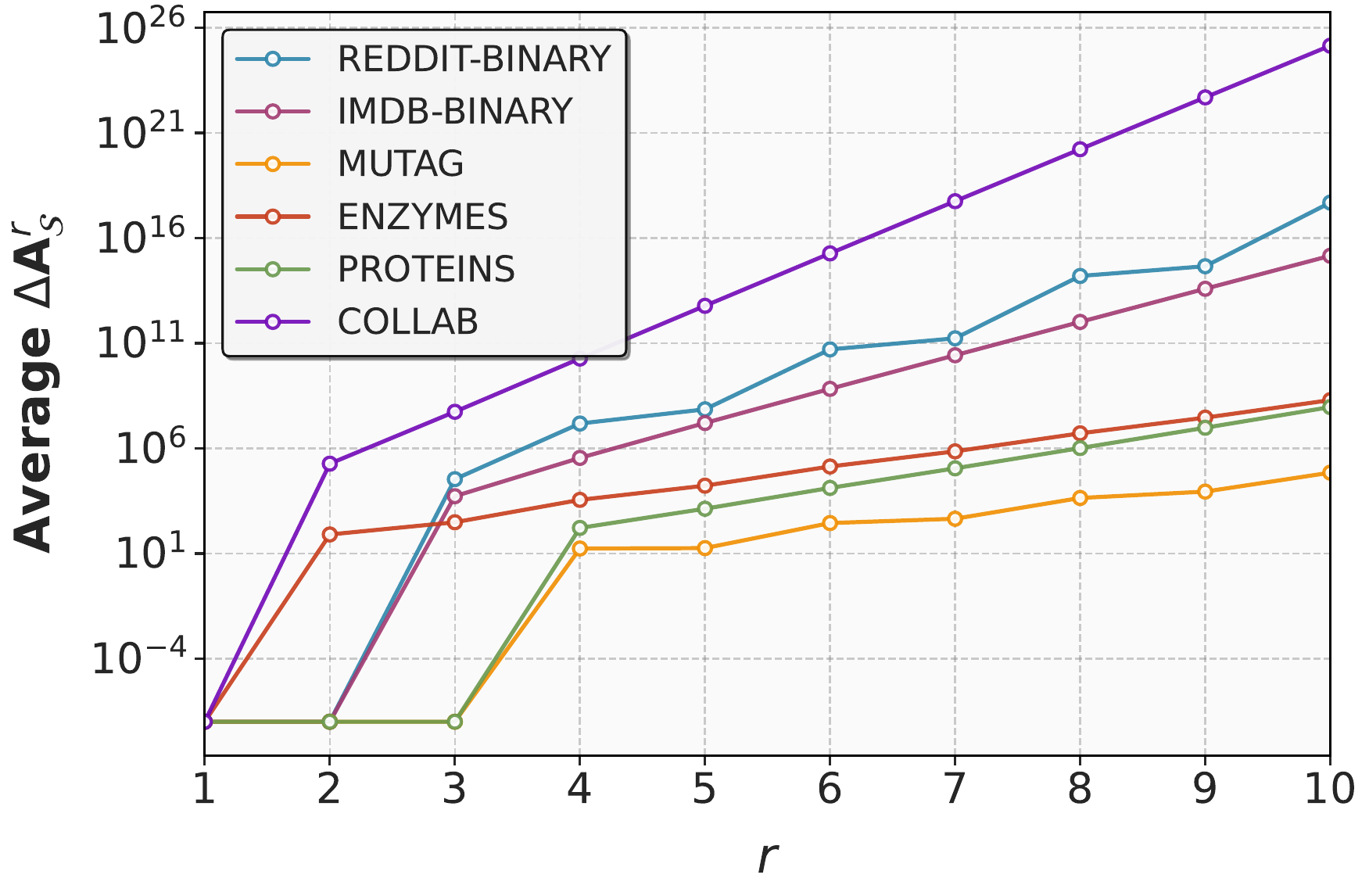}
	\caption{Average change in number of paths with increasing walk length $r$.}
	\label{fig:connectivity_measure}
\end{figure}

To obtain an increase in the total number of walks between non-central nodes when local virtual nodes are added to the graph, we compute
\begin{equation}
	\Delta \mathbf{A}_{\mathcal{S}}^r = \sum_{(i,j) \in \mathcal{S}} (\mathbf{A}^{\prime})^{r}_{ij} - (\mathbf{A})^r_{ij}
\end{equation}
where $\mathbf{A}^{\prime}$ denotes the adjacency matrix of the transformed
graph. We compute the average of $\Delta \mathbf{A}_{\mathcal{S}}^r$ across the
graphs in the dataset.
Figure~\ref{fig:connectivity_measure} shows the average change in the number
of paths as walk length, $r$, increases. The figure indicates that adding LVNs can exponentially increase available paths between nodes beyond a certain walk
length. Thus, we infer that adding LVNs around central regions improves connectivity in the graph.

\subsection{Virtual Node Embedding Analysis}

After demonstrating that LVNs improve structural connectivity, we analyzed the virtual node embeddings to investigate how the proposed approach can exploit the improved
structure by learning rich and diverse representations. We focus on three
important questions to evaluate the effectiveness of virtual node embeddings:
\textbf{Q1)} Can the model learn meaningful LVN embeddings, or do the embeddings merely contribute as noise? \textbf{Q2)} Can the LVN embeddings encode
diverse information?
\textbf{Q3)} Do the LVN embeddings capture structurally rich information?

\subsubsection{Weight Evolution}

\begin{figure*}[t!]
	\centering
	\begin{tabular}{ccc}
		\includegraphics[width=0.28\textwidth]{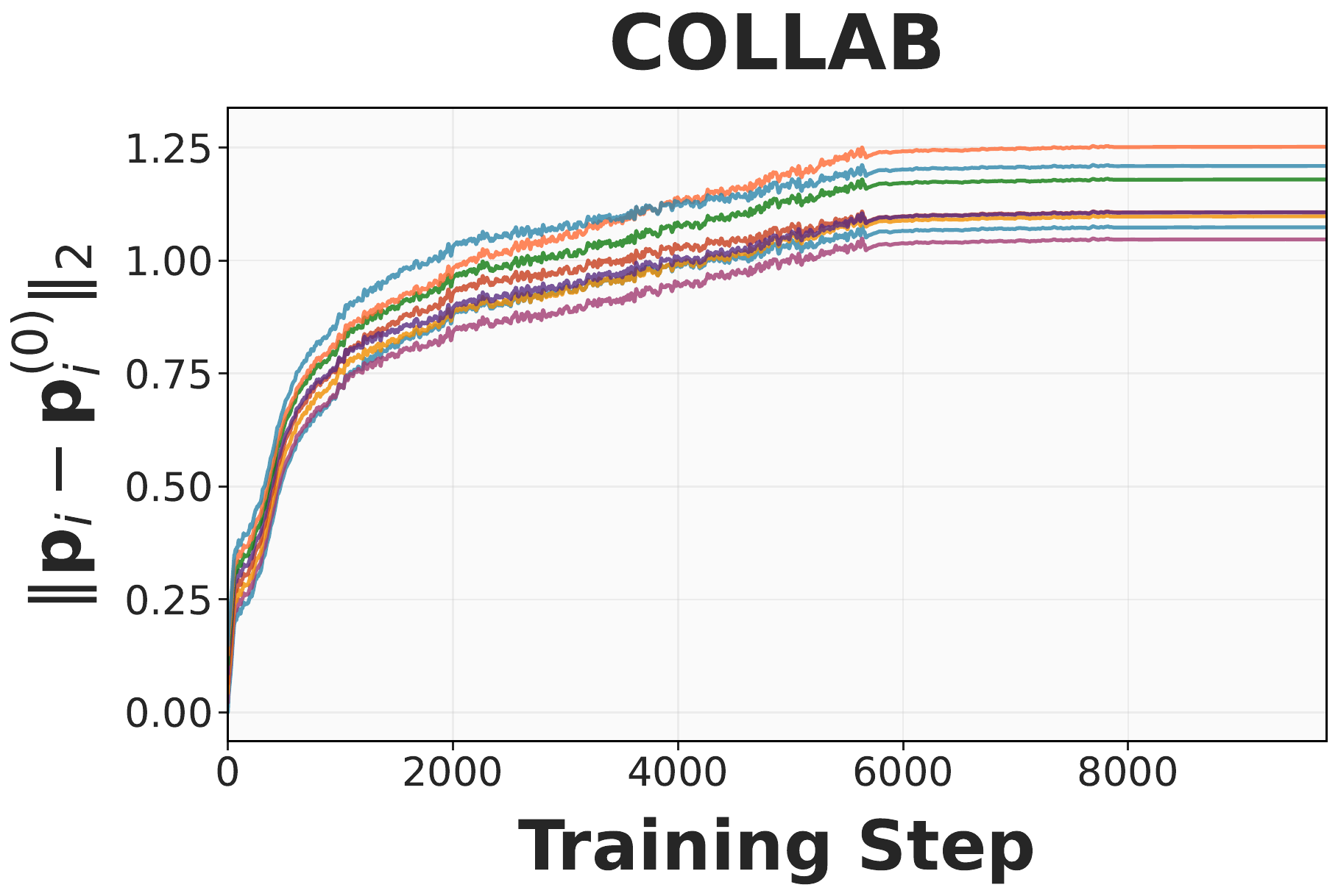}      &
		\includegraphics[width=0.28\textwidth]{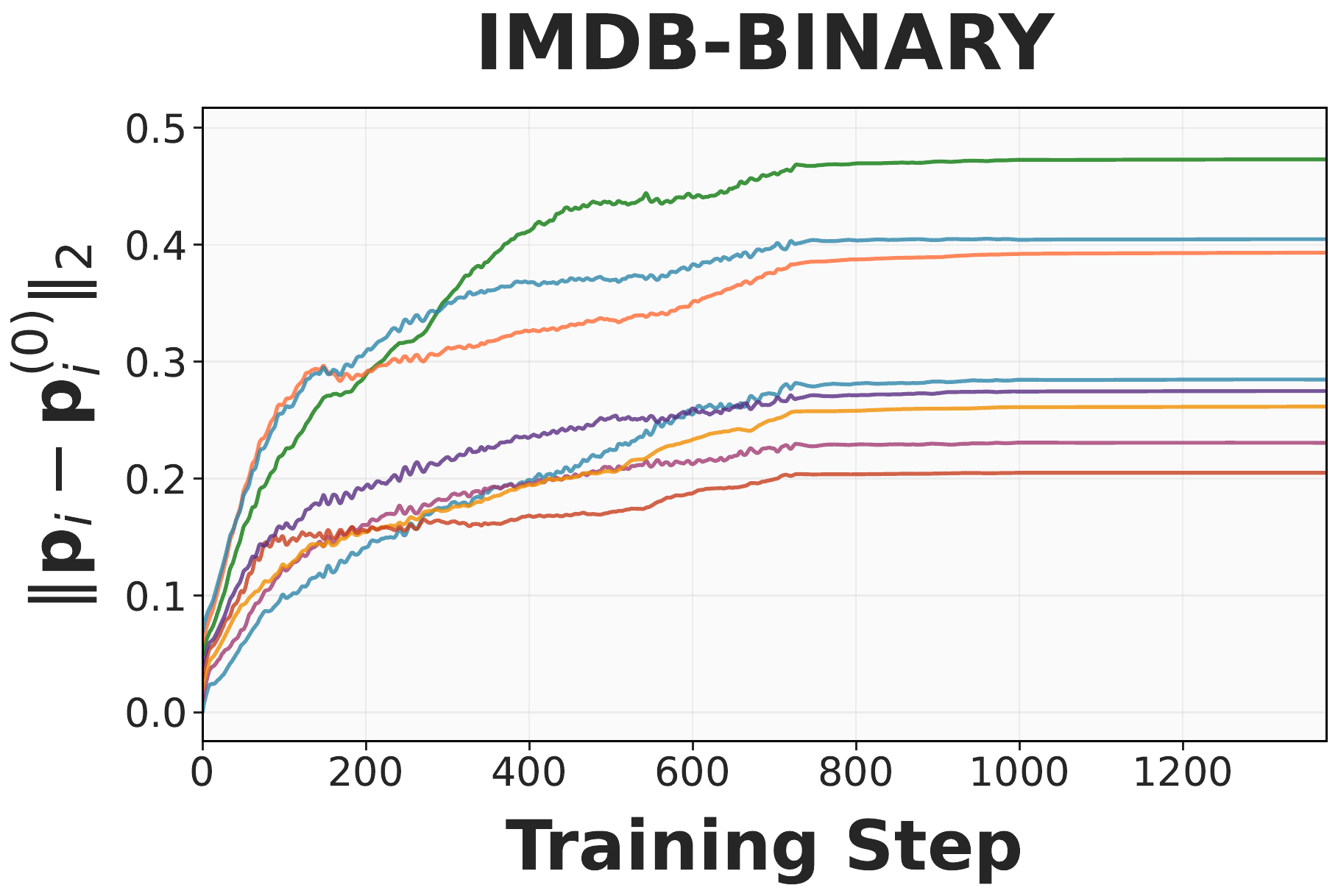} &

		\includegraphics[width=0.28\textwidth]{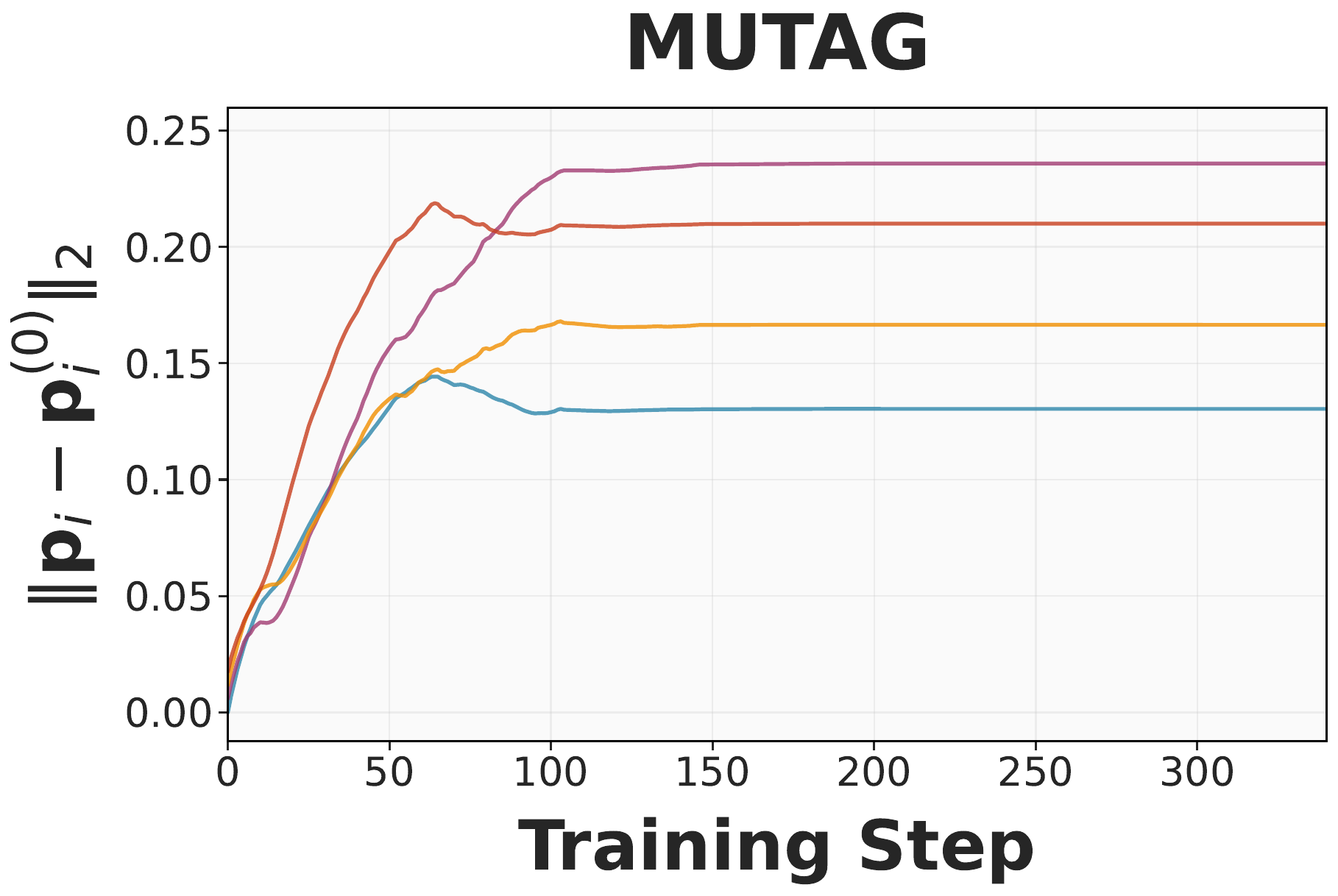}         \\
		\includegraphics[width=0.28\textwidth]{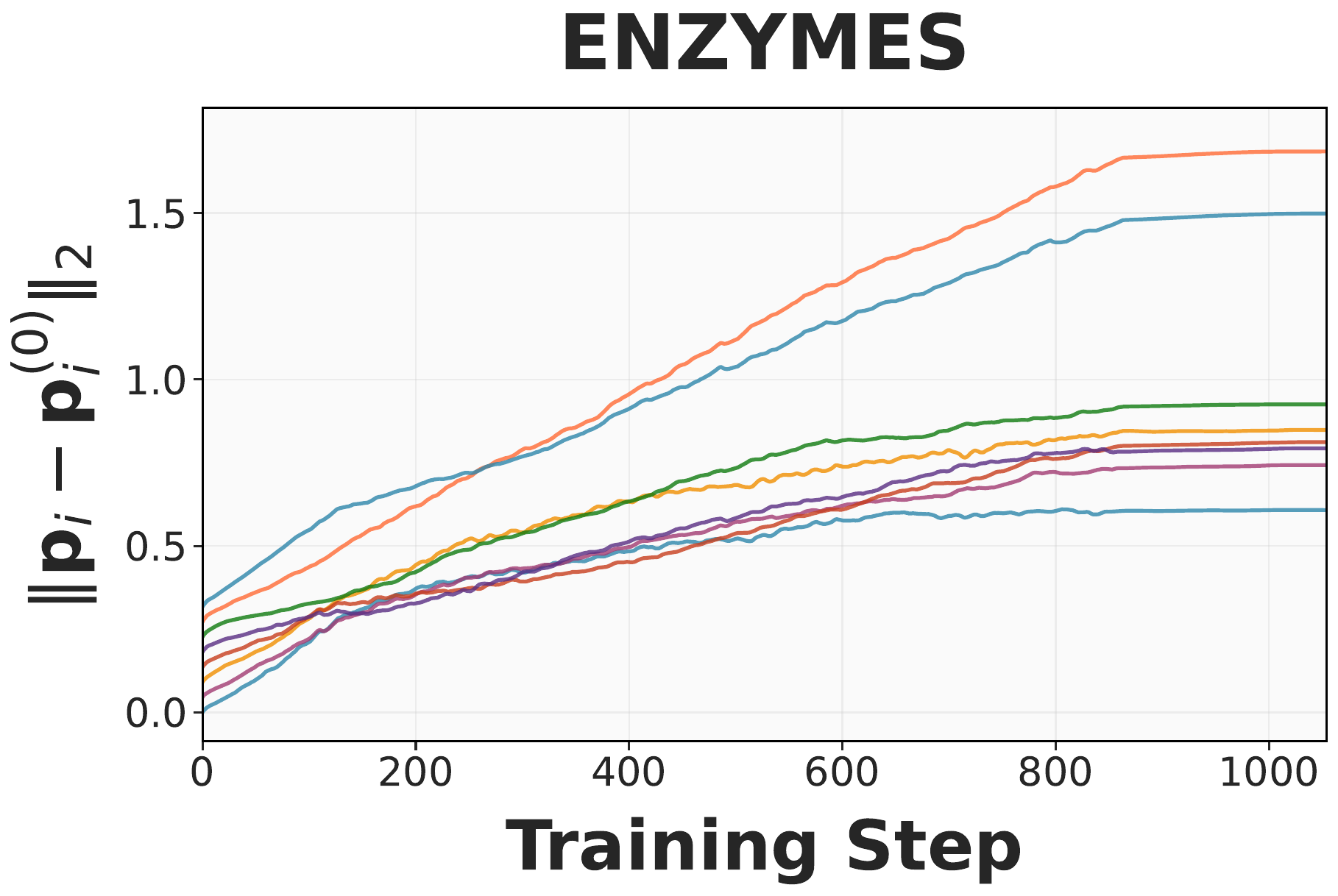}     &

		\includegraphics[width=0.28\textwidth]{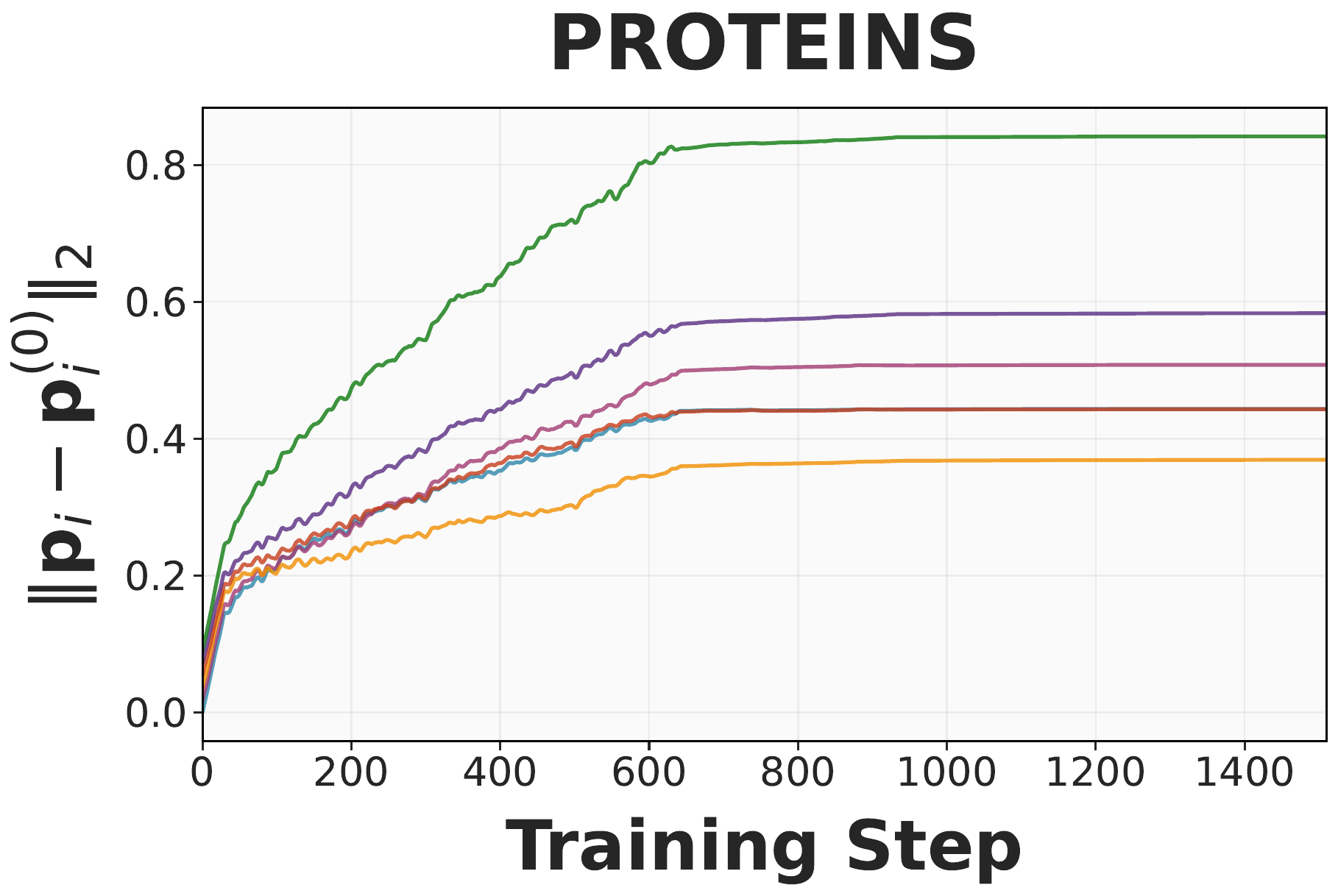}    &
		\includegraphics[width=0.28\textwidth]{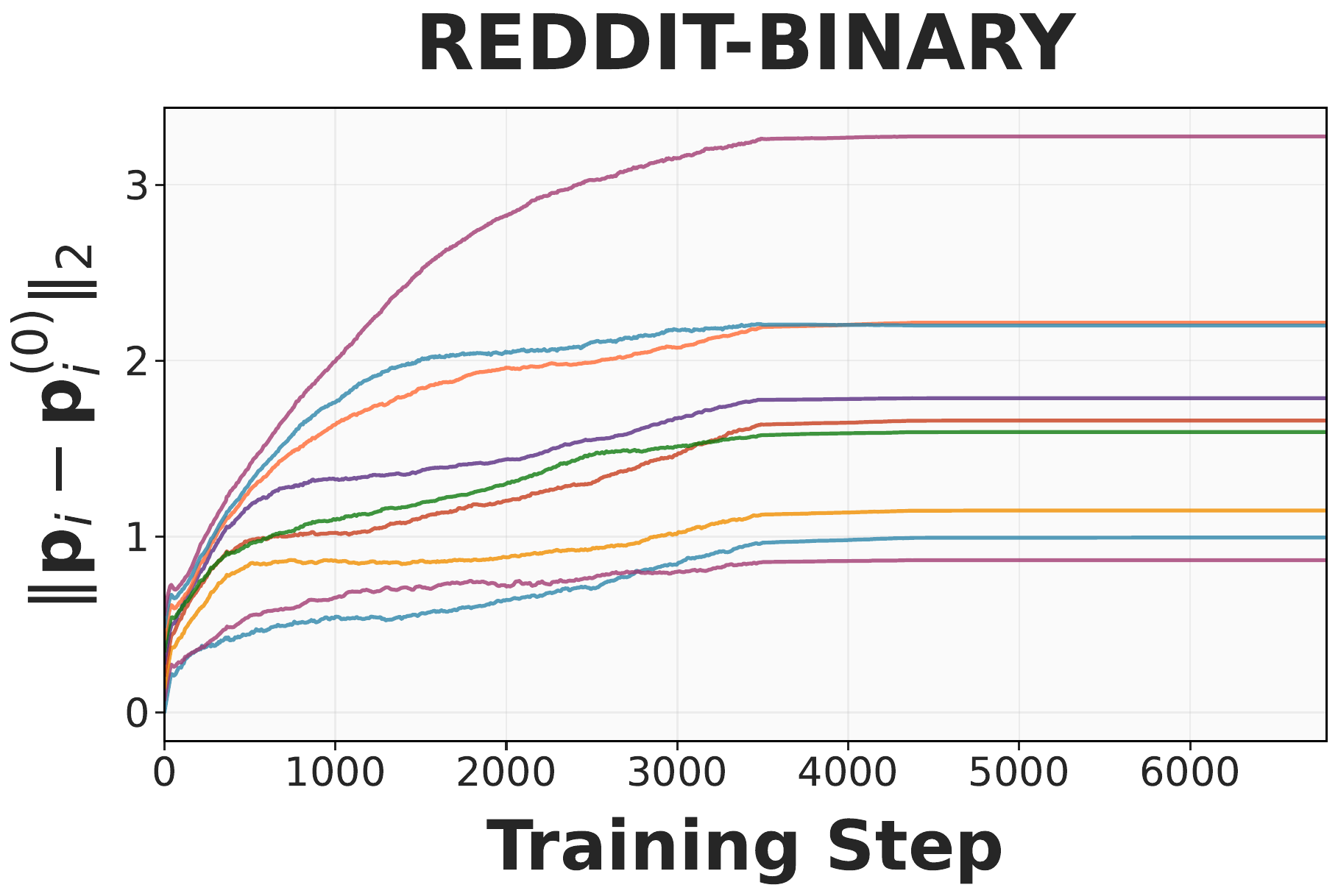} \\
	\end{tabular}
	\caption{The evolution of the embedding weights during training for
		TUDataset~\cite{morris2020tudataset}. The curves in each plot
		represent the weight change of a specific LVN embedding. The number of LVNs
		is tuned.}
	\label{fig:emb_evolution}
\end{figure*}

We investigate how the randomly initialized embeddings evolve during training.
For each embedding $i$, we track the distance $\| \mathbf{p}_i -
	\mathbf{p}^{(0)}_i \|_2$ during training where $\mathbf{p}_i$ represents the
current embedding vector and $\mathbf{p}_i^{(0)}$ denotes the initial embeddiing
value determined randomly. Figure~\ref{fig:emb_evolution} shows that LVN embeddings are indeed updated during training. This result demonstrates that the LVN embeddings do not merely act as random noise to differentiate each virtual node but serve as meaningful representations for the downstream task.

\subsubsection{Mutual Similarity}

\begin{figure*}[t!]
	\centering
	\begin{tabular}{cccccc}
		\includegraphics[width=0.14\textwidth]{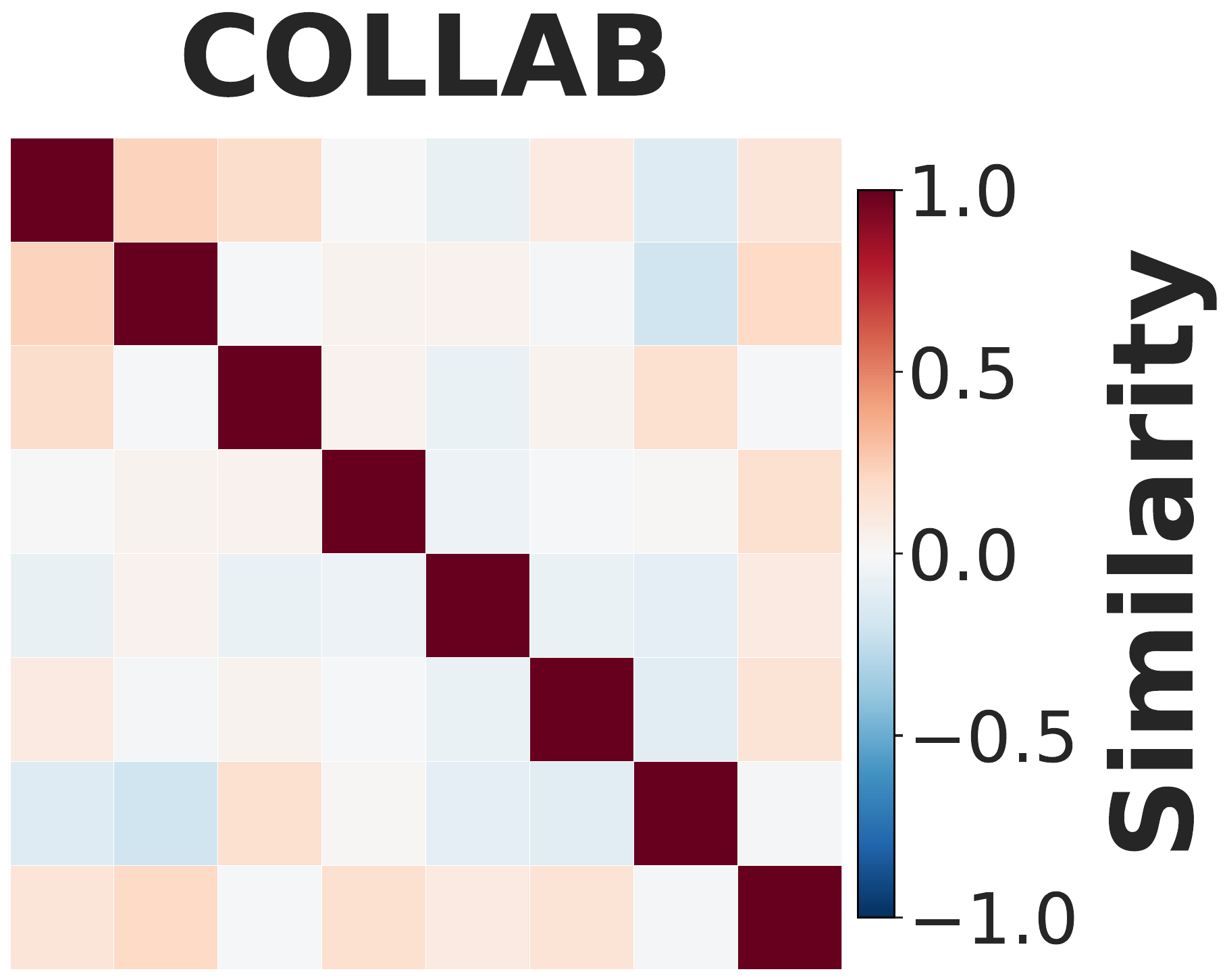}      &
		\includegraphics[width=0.14\textwidth]{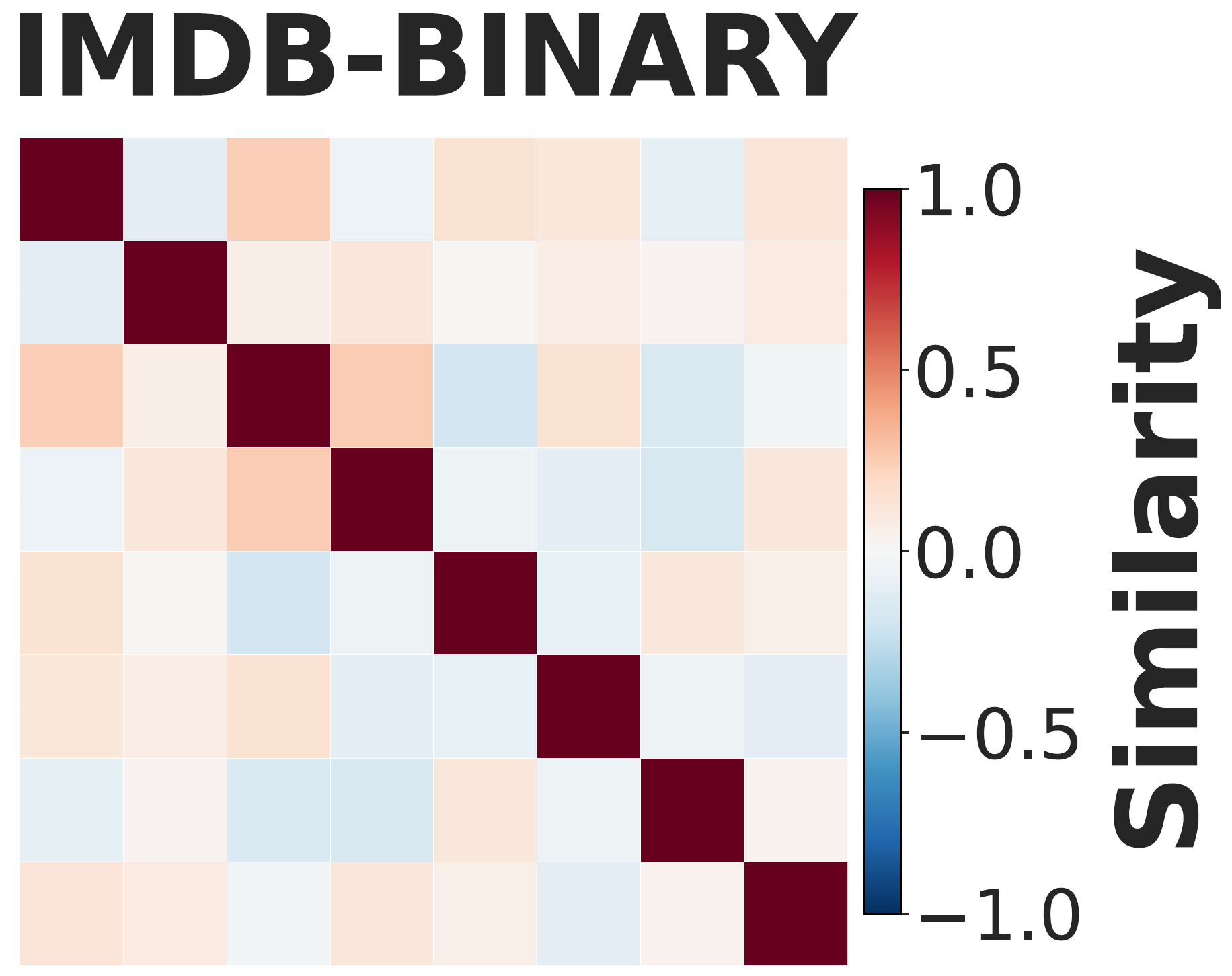} &
		\includegraphics[width=0.14\textwidth]{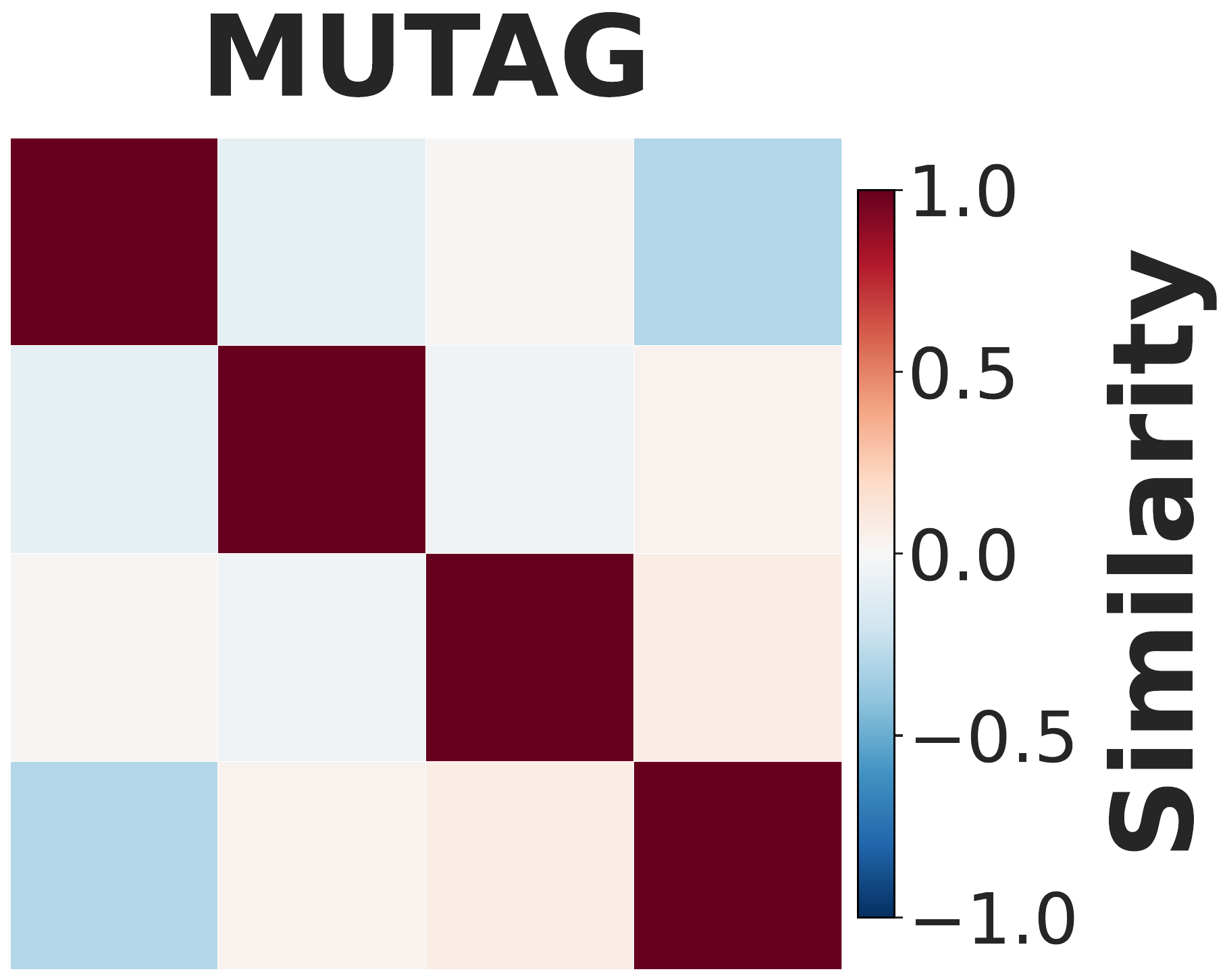}       &
		\includegraphics[width=0.14\textwidth]{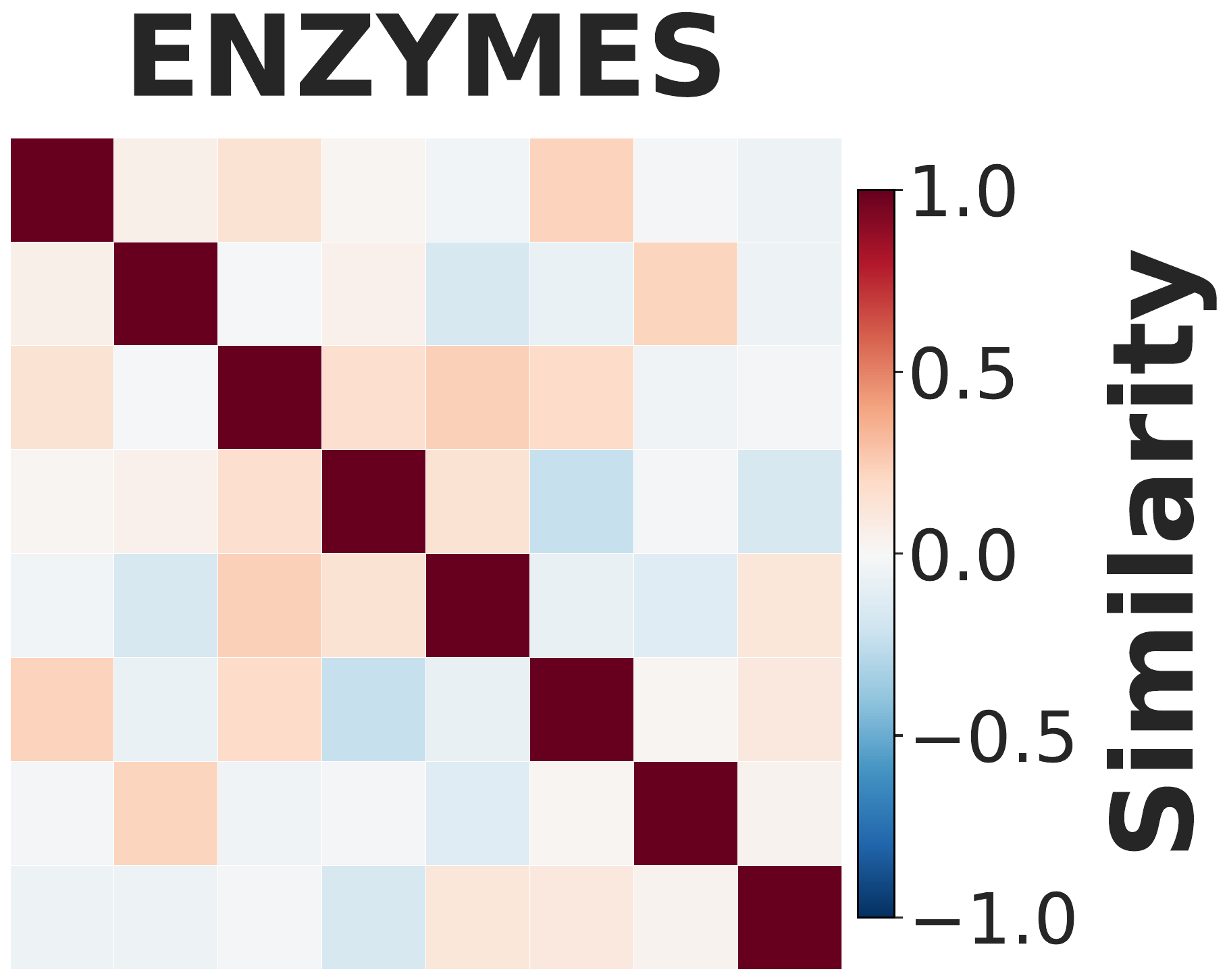}     &
		\includegraphics[width=0.14\textwidth]{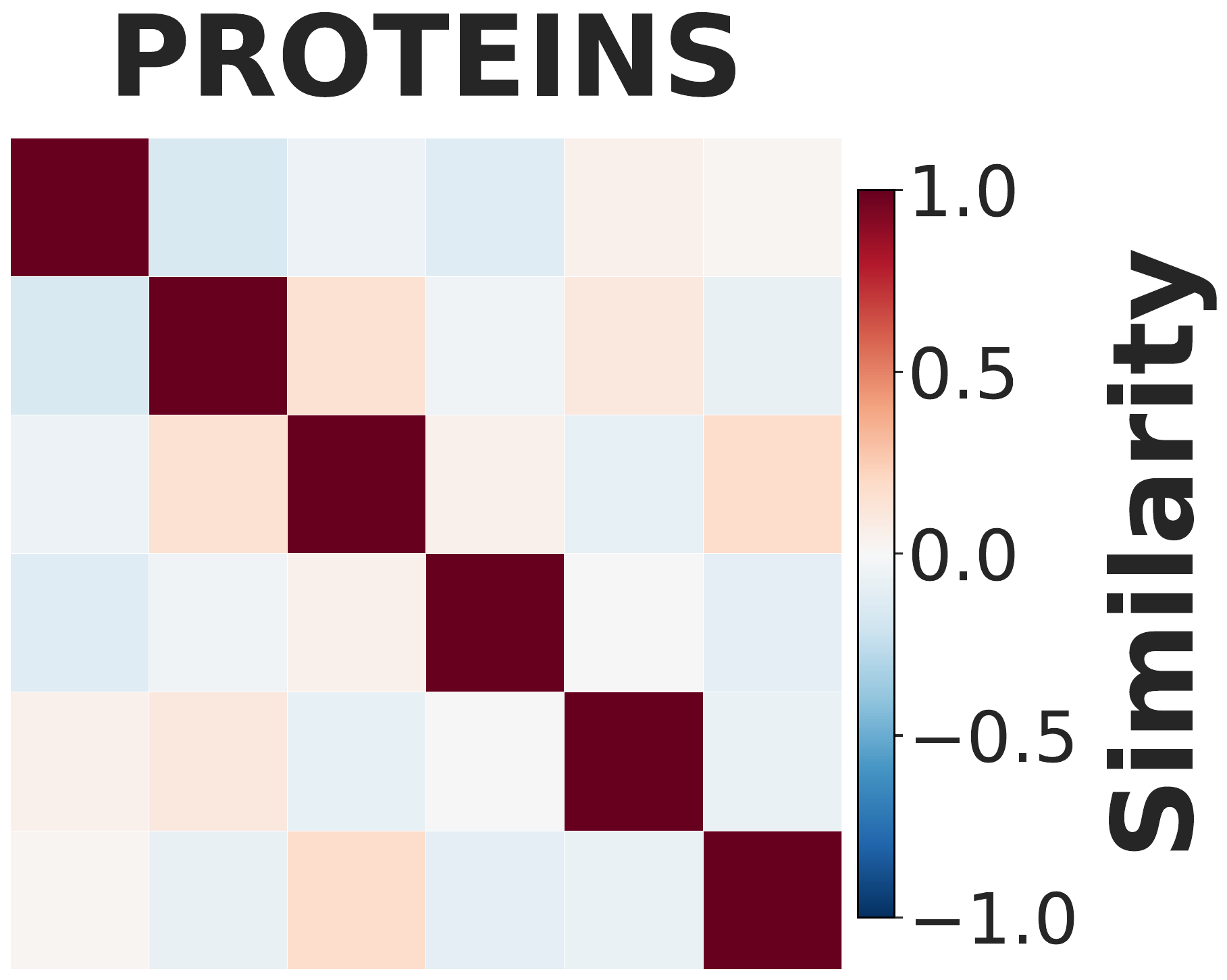}    &
		\includegraphics[width=0.14\textwidth]{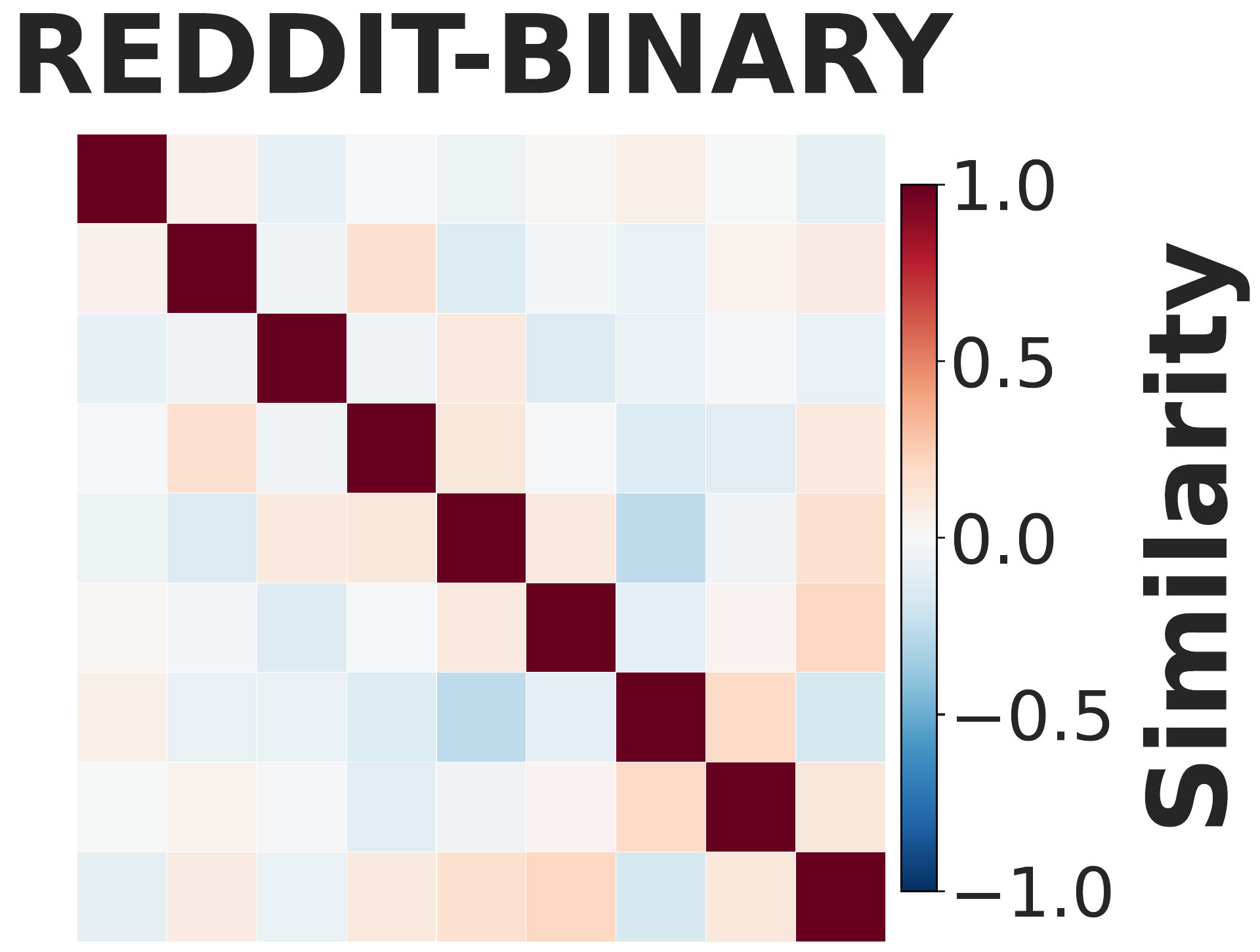} \\
		\includegraphics[width=0.14\textwidth]{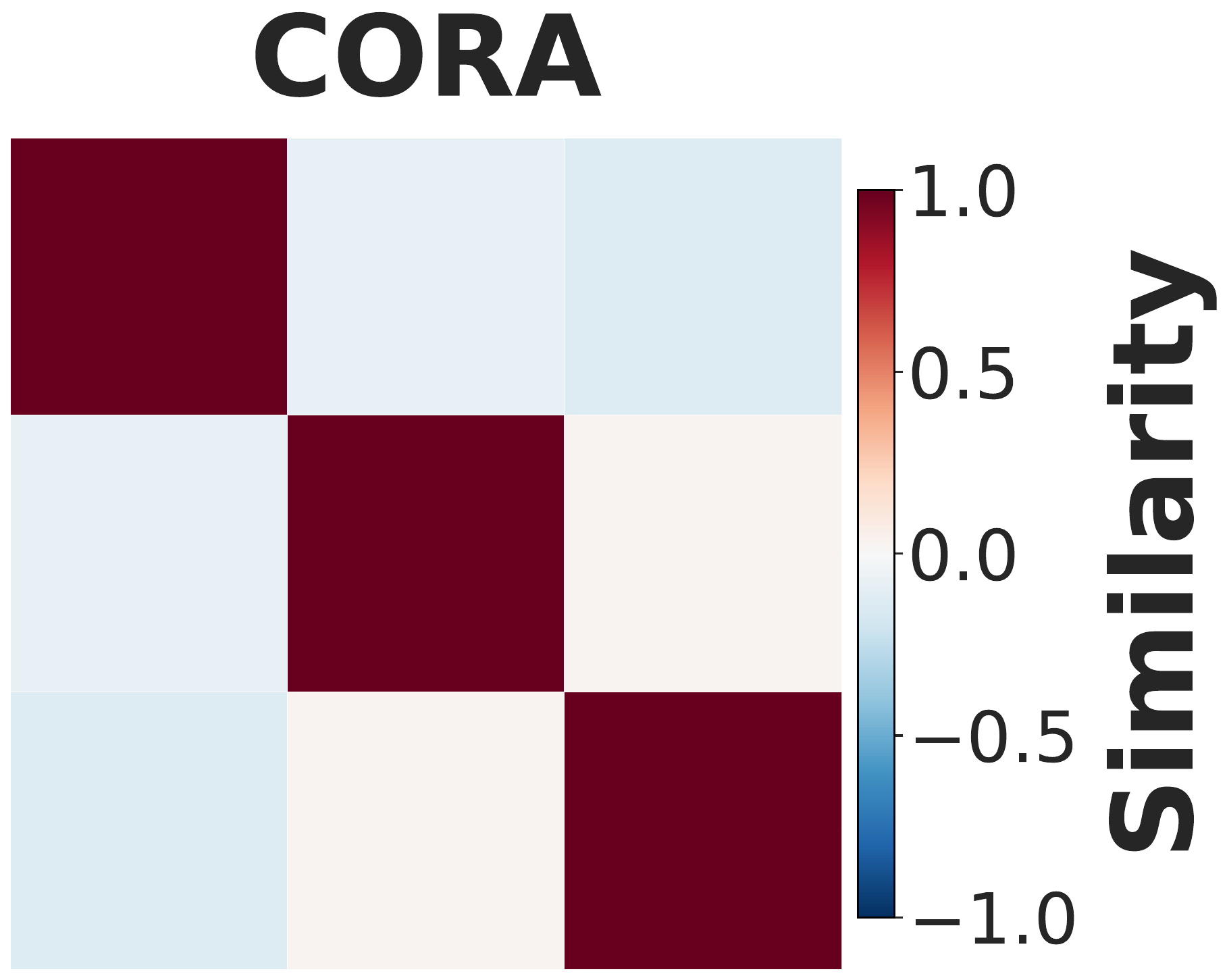}         &
		\includegraphics[width=0.14\textwidth]{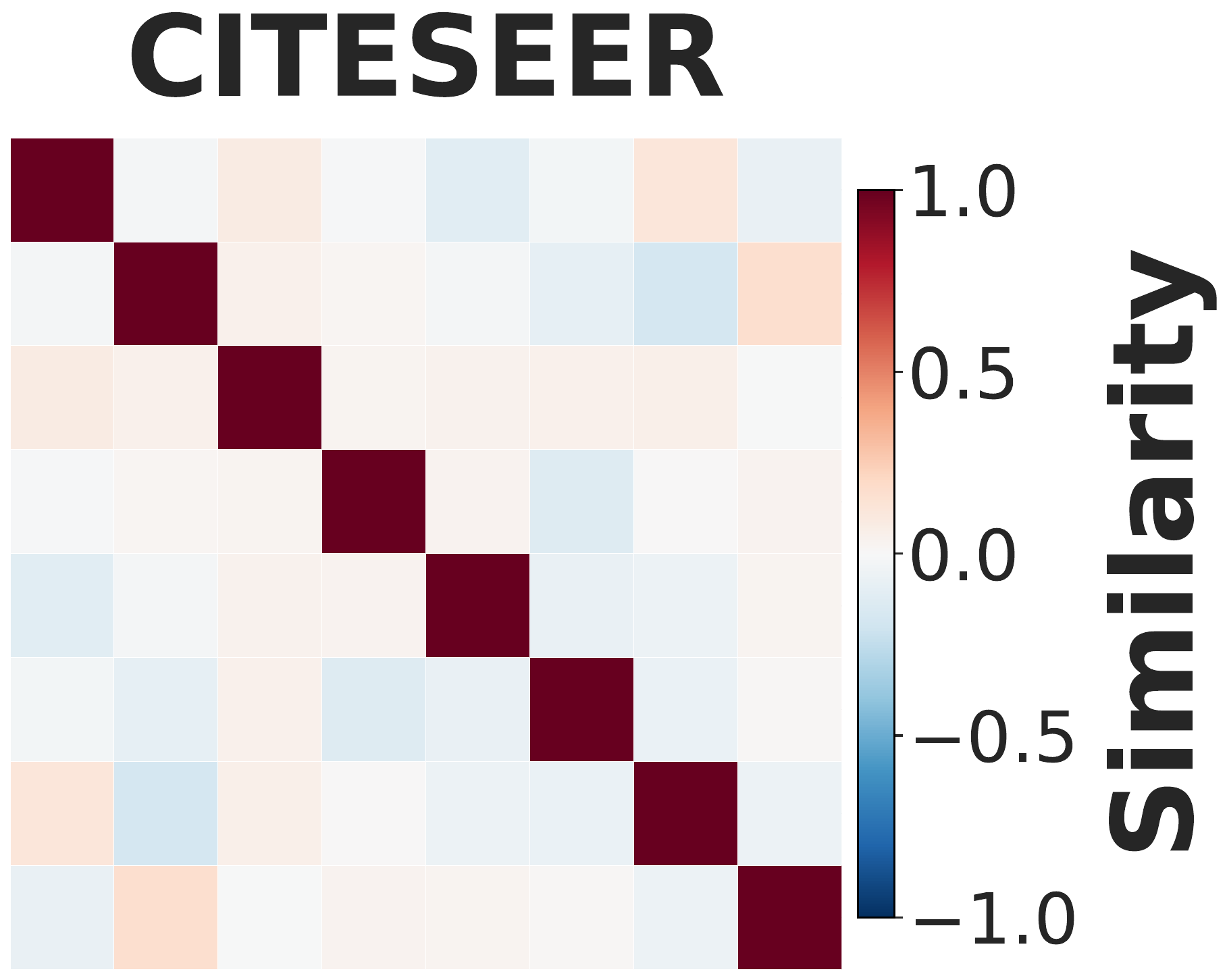}     &
		\includegraphics[width=0.14\textwidth]{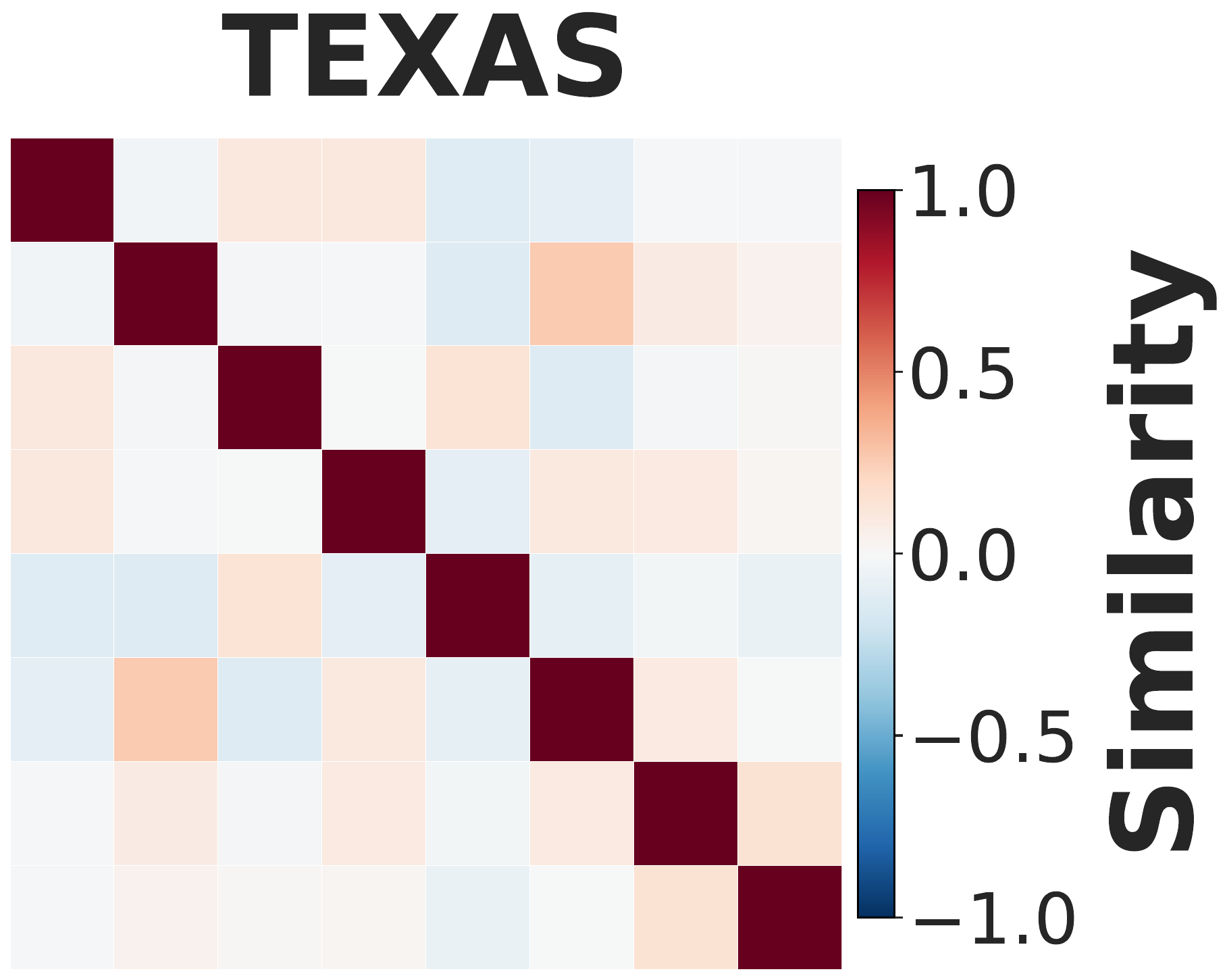}        &
		\includegraphics[width=0.14\textwidth]{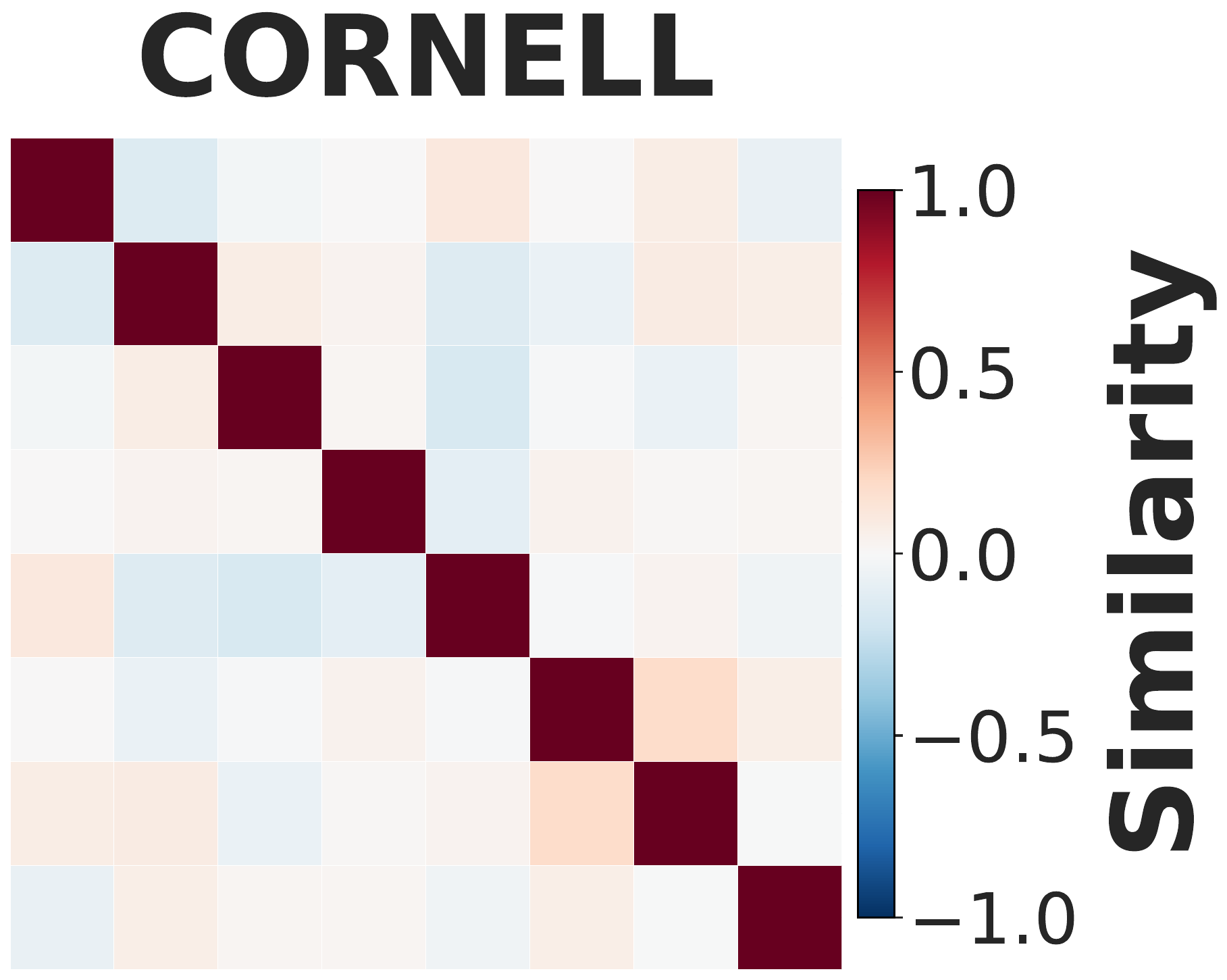}      &
		\includegraphics[width=0.14\textwidth]{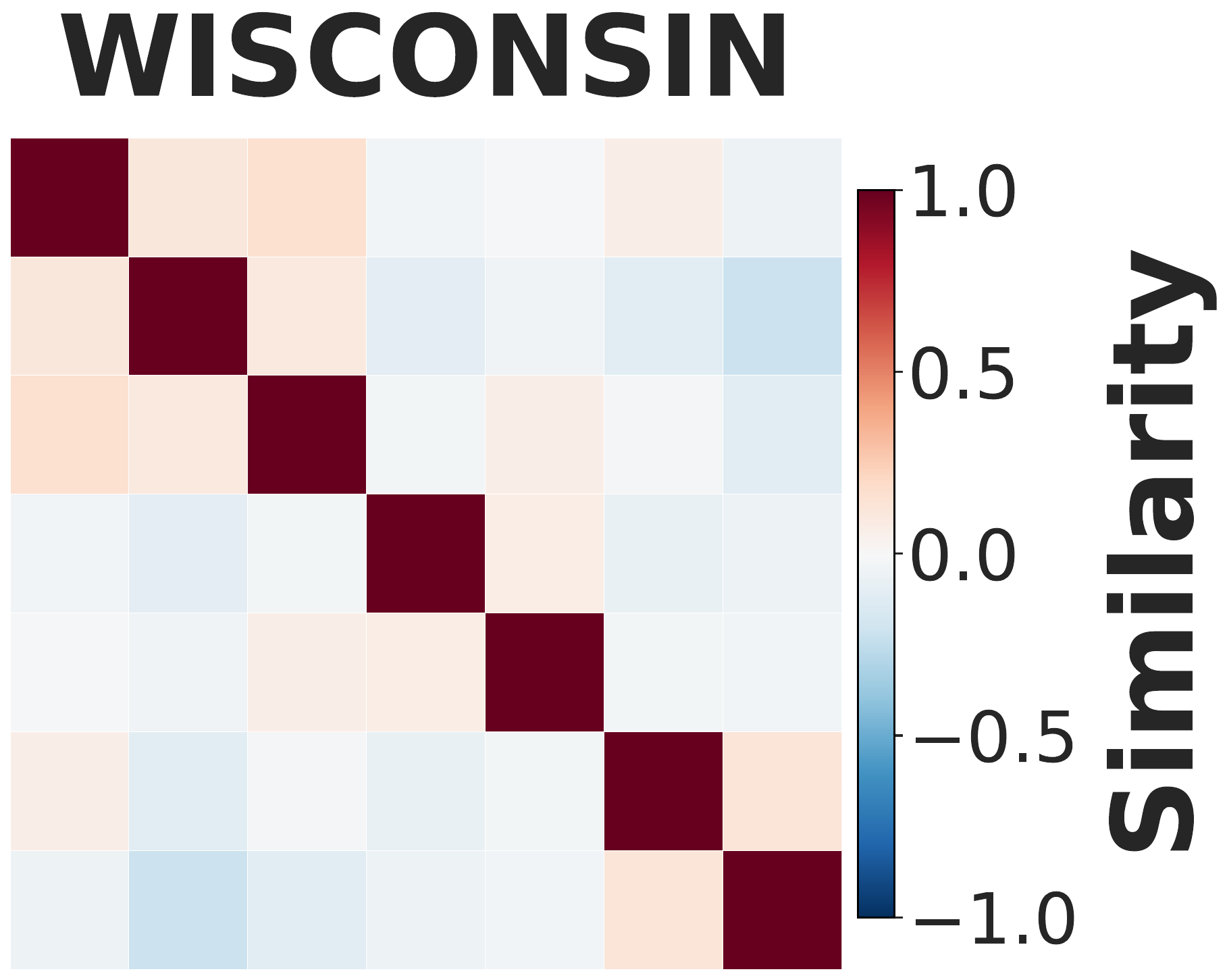}    &
		\includegraphics[width=0.14\textwidth]{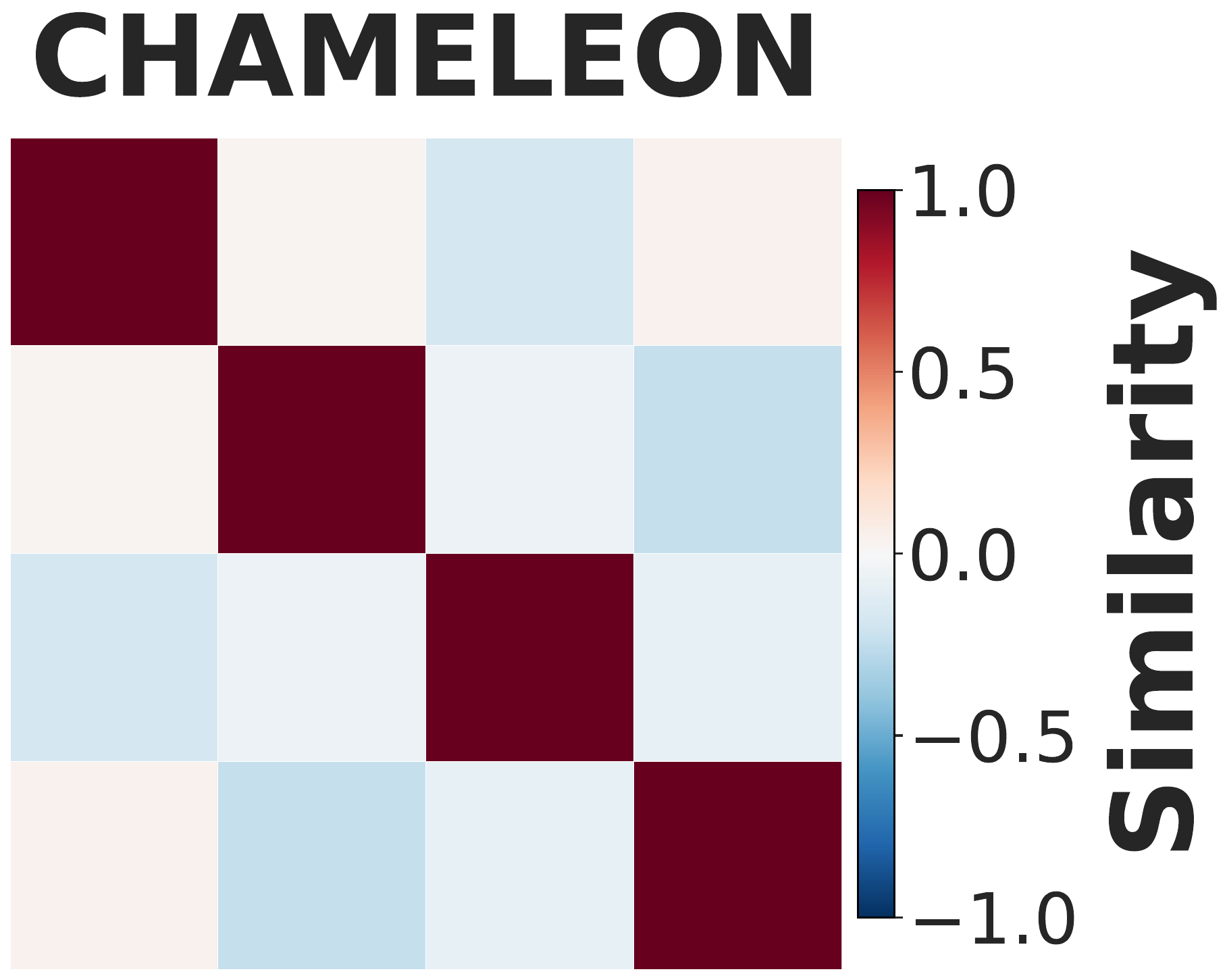}      \\
	\end{tabular}
	\caption{LVN embedding heatmaps for graph classification datasets (top row) and node classification datasets (bottom row). The number of LVNs is tuned.}
	\label{fig:emb_heatmaps}
\end{figure*}

We demonstrated that each LVN embedding is updated and converges to a specific value during training. Another desired property is that learned
virtual node embeddings should represent diverse and complementary information rather than redundant, nearly identical features. Therefore, we visualized the mutual similarity between trained virtual node embeddings for
each task and dataset. Figure~\ref{fig:emb_heatmaps} demonstrates that LVN embeddings consistently differ. We observe that the
embedding cosine similarities are usually around zero, while some pairs of embeddings may have similarity close to $1$ or $-1$. Therefore, the similarities
show that the source of embedding distinctiveness is not random initialization but their utility in the learning process. In conclusion, the LVN embeddings allow us to exploit the additional
representational capacity introduced by the LVN groups.

\subsubsection{Discriminative Power}

We finally investigated how much domain and structural information LVN embeddings encode during training. For this analysis, we compared the performance of the graph classification task
using only the input features $\mathbf{x}_v^{(0)}$ for $v \in \mathcal{V}$ with
and without virtual node embeddings. We trained a multi-layer perceptron (MLP) with two layers that processes each node's
features without using the graph structure and predicts the output with global
mean pooling across all nodes in the graph. To assess the contribution of
virtual node embeddings, we compare MLP performance using input features from
the raw graph versus input features from graphs equipped with LVNs and
their embeddings. We use pre-trained LVN embeddings as the MLP's input for the second case. To prevent data leakage, we store the original 50 splits and train the MLPs on each split with its corresponding pre-trained embeddings. Table~\ref{tab:embedding_ablation} displays the graph classification performance
of MLPs trained with only input features versus those trained with input
features and virtual node embeddings. We only report results on graph classification datasets with input node features to ensure fair comparison. The results show that MLPs trained with virtual node embeddings nearly reach the
performance of GCNs trained with local virtual nodes. The MLP performance almost
matches the GCN's on MUTAG while remaining competitive on ENZYMES and
PROTEINS. Thus, we infer that LVN embeddings can capture valuable structural information during training. The performance difference between MLPs trained with raw input
features (structureless baseline) and those trained with additional pre-trained LVN embeddings reveals how much structural information the virtual node
embeddings contribute to the learning process.

\begin{table}[t!]
	\centering
	\caption{Comparison of the graph classification performance of MLPs trained with only the original node input features and with LVN embeddings added. Graph datasets without node features, REDDIT-BINARY, IMDB-BINARY, and COLLAB, are excluded.}
	\resizebox{\linewidth}{!}{
		\setlength{\tabcolsep}{4pt}
		\begin{tabular}{lcc}
			\toprule
			\textbf{Dataset} & \textbf{Input Features} & \textbf{+Virtual Node Embeddings} \\
			\midrule
			MUTAG            & $68.000 \pm 2.947$      & $\mathbf{84.000 \pm 2.867}$       \\
			ENZYMES          & $27.933 \pm 1.682$      & $\mathbf{29.567 \pm 1.803}$       \\
			PROTEINS         & $68.511 \pm 1.701$      & $\mathbf{73.532 \pm 1.906}$       \\
			\bottomrule
		\end{tabular}
	}
	\label{tab:embedding_ablation}
\end{table}

\section{Conclusion}

This study proposes to increase graph representation capacity and connectivity by augmenting graphs with local virtual
nodes (LVNs). The LVNs can alleviate the negative impact of bottlenecks on GNN training by creating additional pathways for information flow when they are added to nodes with high centrality. We assign trainable embeddings to LVNs, but they share embeddings across different central regions of the graph. Thus, in addition to integrating additional pathways by adding LVNs, long-range communication during message-passing is facilitated by trainable shared LVN embeddings without the need for adding extra GNN layers. Unlike graph rewiring and global virtual node techniques that tackle over-squashing by altering graph topology, we avoid modifying the global graph structure or removing original edges. The experimental analyses report that the proposed approach can outperform state-of-the-art baselines in graph and node classification benchmarks. Moreover, we empirically show that adding LVNs can improve the connectivity between nodes in terms of effective resistance. On the other hand, the proposed approach may struggle in settings where high-quality input node
features are present and play a crucial role in solving the task as much as the
graph structure. In such cases, more efficient feature initialization approaches
are necessary to ensure the virtual node group effectively represents the input
representation of their corresponding central nodes while avoiding
over-smoothing. Lastly, determining the set of nodes to replace with virtual node groups is a crucial part of our algorithm. The type of
nodes that could benefit from LVNs may vary based on domain and task. In future work, we
aim to replace this step with learnable modules that automatically select the nodes to
expand based on graph structure and task requirements rather than centrality-based heuristics.




\bibliographystyle{IEEEtran}
\bibliography{references}






\vfill

\end{document}